\documentclass[11pt]{article}

\usepackage{acl}

\usepackage{times}
\usepackage{latexsym}
\usepackage[T1]{fontenc}
\usepackage[utf8]{inputenc}
\usepackage{microtype}
\usepackage{inconsolata}
\usepackage{graphicx}
\usepackage{booktabs}
\usepackage{amsmath}
\usepackage{amssymb}
\usepackage{multirow}
\usepackage{needspace}
\usepackage{listings}
\usepackage{mathtools}
\usepackage{colortbl}
\usepackage{threeparttable}
\usepackage{tabularx}
\usepackage{placeins}
\usepackage{dblfloatfix}
\usepackage{balance}
\usepackage[ruled,vlined,linesnumbered]{algorithm2e}

\newcolumntype{Y}{>{\raggedright\arraybackslash}X}
\AtBeginDocument{%
  \setlength{\abovedisplayskip}{6pt plus 2pt minus 2pt}%
  \setlength{\belowdisplayskip}{6pt plus 2pt minus 2pt}%
  \setlength{\abovedisplayshortskip}{4pt plus 2pt minus 1pt}%
  \setlength{\belowdisplayshortskip}{4pt plus 2pt minus 1pt}%
}

\title{
How Do Document Parsers Break?\\
Auditing Structural Vulnerability in Document Intelligence
}

\author{
\textbf{Yue Chen\textsuperscript{1,*}},
\textbf{Yihao Wang\textsuperscript{1,*}},
\textbf{Ziyi Tang\textsuperscript{1,3}},
\textbf{Yongsen Zheng\textsuperscript{2}},
\textbf{Keze Wang\textsuperscript{1,\textdagger}}
\\
\\[-2pt]
\textsuperscript{1}Sun Yat-sen University \\
\textsuperscript{2}Nanyang Technological University \\
\textsuperscript{3}Chaotic Pendulum Lab \\
\\
{\fontsize{10.95pt}{12.326pt}\selectfont
\texttt{\{cheny2639,wangyh357,tangzy27\}@mail2.sysu.edu.cn},
}
\\
{\small
\texttt{yongsen.zheng@ntu.edu.sg}, \quad
\texttt{kezewang@gmail.com}
}
\\
{\small
\textsuperscript{*}Equal contribution.
\quad
\textsuperscript{\textdagger}Correspondence:
\href{mailto:kezewang@gmail.com}{kezewang@gmail.com}
}
\\
{\small
\textbf{GitHub Repository:} \url{https://github.com/ef1026/ProSA}
}
}

\begin{document}
\maketitle

\begin{abstract}
\fussy
\setlength{\emergencystretch}{2em}
Document Layout Analysis (DLA) pipelines provide structured page representations for retrieval-augmented generation, long-document question answering, and related applications. Yet their robustness evaluation remains largely area-centric.
We identify this \textit{Footprint Bias} and propose \textbf{ProSA}, a lightweight output-level auditing framework that decouples controlled probing, policy-driven targeting, and structure-aware diagnosis.
ProSA combines Block-level Structural Loss Rate (B-SLR), granularity-aware exposure descriptors, and pathway attribution to analyze where structural identity is lost, at what exposure granularity failures emerge, and how failures propagate.
Across MinerU and PP-StructureV3 on 1{,}000 pages, affected area weakly tracks perturbation-induced OCR instability ($R^2{=}0.384/0.110$), whereas B-SLR aligns much more closely with it ($R^2{=}0.727/0.916$).
Exposure descriptors further separate occlusion- and topology-dominant pathways, while matched-footprint structural probes cause much larger downstream QA/retrieval drops than area-matched erasure.
These results shift DLA robustness evaluation from footprint-based measurement toward structure-aware vulnerability auditing.
\end{abstract}

\section{Introduction}
\label{sec:intro}

Document Layout Analysis (DLA) converts visually organized pages into structured representations of text, tables, figures, and other layout elements, supporting document intelligence systems such as retrieval-augmented generation over visually rich documents \citep{ueda-etal-2026-scan}, financial document reasoning \citep{zhao-etal-2024-docmath}, and medical record understanding \citep{cottet-etal-2026-lightweight}.
Errors at this stage---including boundary corruption, spurious block merges, and misplaced outputs---can propagate as corrupted context, misaligned evidence, and unreliable downstream predictions.

Despite progress on clean benchmarks and realistic evaluation settings, DLA robustness evaluation remains largely area-centric.
Existing protocols often parameterize perturbation severity by global corruption magnitude or affected pixel footprint, and report aggregate degradation through metrics such as Character Error Rate (CER) or detection accuracy \citep{hendrycks2019,michaelis2020,rodla2024,docptbench2025,Ouyang_2025_CVPR}.
Such evaluations tell us \emph{whether} performance declines, but reveal little about \emph{how} or \emph{why} parsing fails structurally, especially when lightweight auditing is needed on unlabeled document collections.

\begin{figure}[t]
  \centering
  \includegraphics[width=\linewidth]{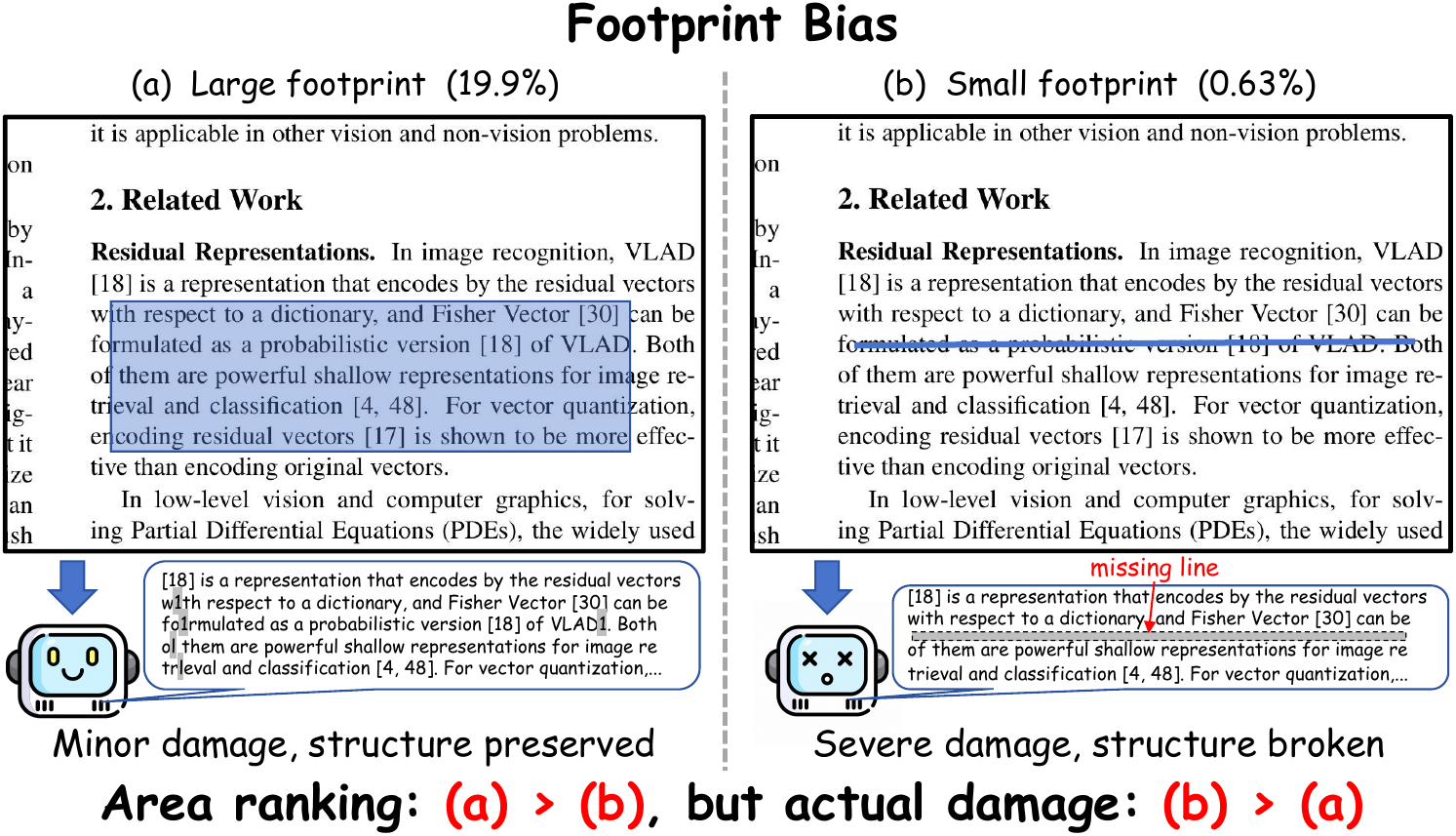}
\caption{\textbf{Footprint Bias}: a large perturbation may preserve structure, while a small structural probe can drop a text line and cause greater parsing damage.}
\label{fig:motivation}
\end{figure}

A central limitation is what we term the \textbf{Footprint Bias}: the tendency to infer perturbation severity from pixel footprint alone.
As Figure~\ref{fig:motivation} shows, a broad erased region may corrupt local text while leaving the main paragraph structure largely intact.
By contrast, a thin perturbation with a much smaller footprint can intersect a sensitive line or layout boundary, causing a missing-line failure in which the parser skips a line and connects non-adjacent text fragments as if they were continuous.
The smaller perturbation can therefore be more damaging: it removes document evidence and disrupts the local reading flow despite occupying far fewer pixels.

Such mismatches can propagate downstream: a dropped line may remove answer-bearing evidence or corrupt retrieval/QA context, while a larger visible perturbation may leave block identity and reading order intact.
They also expose why area-centric evaluation can mis-rank structural vulnerability.
The key challenge is therefore not simply to measure degradation, but to explain structural vulnerability through three questions:
(Q$_1$) where structural identity is lost,
(Q$_2$) at what granularity perturbation exposure becomes predictive of failure, and
(Q$_3$) which pathway---direct physical occlusion or topology-level disruption---drives the degradation.

To address this diagnostic problem, we propose \textbf{ProSA}
(\textbf{Pro}be-guided \textbf{S}tructure-aware \textbf{A}uditing),
a lightweight output-level framework for auditing structural vulnerability in DLA pipelines.
ProSA instantiates the audit as a tripartite decomposition
$\mathcal{F}=(\mathcal{S},\Pi,\mathcal{D})$, where $\mathcal{S}$ defines a shared space of controlled visual probes, $\Pi$ selects probes under different targeting policies, and $\mathcal{D}$ diagnoses the resulting clean--perturbed parsing divergence.
By decoupling the probe space, selection logic, and diagnostic criteria, ProSA compares different targeting policies under a common action space without modifying the audited parser.

The diagnostic component is centered on B-SLR, an output-level signal that measures structural loss by comparing clean and perturbed parser outputs.
Unlike footprint-based severity or terminal-only degradation scores, B-SLR checks whether each clean block preserves a valid geometry--text counterpart, capturing both missing geometry and lost textual identity under a unified structural criterion.
Exposure descriptors and pathway attribution then explain at which exposure granularity these losses emerge and whether they arise from direct occlusion or topology-dominant disruption.
Because the core diagnostics use the clean parse as structural reference, ProSA supports lightweight auditing on unlabeled documents while still incorporating benchmark annotations when available.
Experiments show that these structural diagnostics better track OCR instability and downstream QA/retrieval drops than footprint-based severity.

In summary, our contributions are threefold:

\textbf{(1)} We identify the \textbf{Footprint Bias} in DLA robustness evaluation, showing that pixel footprint is an unreliable proxy for structural damage and may mis-rank parser vulnerability.

\textbf{(2)} We propose \textbf{ProSA}, a lightweight output-level auditing framework that couples controlled probes, policy-guided selection, and structure-aware diagnosis.

\textbf{(3)} We instantiate ProSA with B-SLR, exposure descriptors, and pathway attribution, enabling annotation-light analysis that better tracks OCR and QA/retrieval drops.

\section{Related Work}

\paragraph{Document parsing robustness and downstream document intelligence.}
Robustness evaluation has moved from generic corruption suites and detection benchmarks \citep{hendrycks2019,michaelis2020} to document-specific parsing settings such as RoDLA, DocPTBench, and OmniDocBench \citep{rodla2024,docptbench2025,Ouyang_2025_CVPR}. Recent document intelligence studies further show that OCR and layout quality can affect retrieval-augmented generation, financial reasoning, and medical document understanding \citep{ueda-etal-2026-scan,zhao-etal-2024-docmath,cottet-etal-2026-lightweight,ocrhindersrag2025}. However, these evaluations mainly report corruption-level, benchmark-level, or downstream degradation. They provide limited diagnosis of which parser-level structural changes, such as block loss, text-geometry mismatch, or topology disruption, actually drive the observed failures. Our work complements this line by auditing structural failure patterns directly in DLA outputs and relating them to downstream QA/retrieval behavior.

\paragraph{Perturbation footprint, placement, and black-box diagnosis.}
Visual perturbations are often parameterized by magnitude, affected area, or patch budget. Cutout and Random Erasing use erased area as a key augmentation strength \citep{cutout2017,randomerasing2020}, while document augmentation tools expose visual and spatial perturbation parameters for document images \citep{augraphy2023}. Patch-based robustness studies further show that location and composition matter: adversarial patches can remain effective under small spatial budgets \citep{brown2017adversarialpatch}, and recent analyses reveal placement hot-spots or jointly optimize patch shape, location, number, and content \citep{kimhi2026,impact2025}. Black-box explanation methods similarly perturb inputs to identify influential features or sensitive regions \citep{10.1145/2939672.2939778,NIPS2017_8a20a862,petsiuk2018riserandomizedinputsampling}. Unlike these works, we audit structured DLA outputs rather than scalar predictions: each parsed element couples geometry, category, and text. This requires element matching, geometry-text preservation analysis, and occlusion/topology pathway diagnosis, while still requiring no access to parser internals.

\begin{figure*}[!t]
  \centering
  \includegraphics[width=\textwidth]{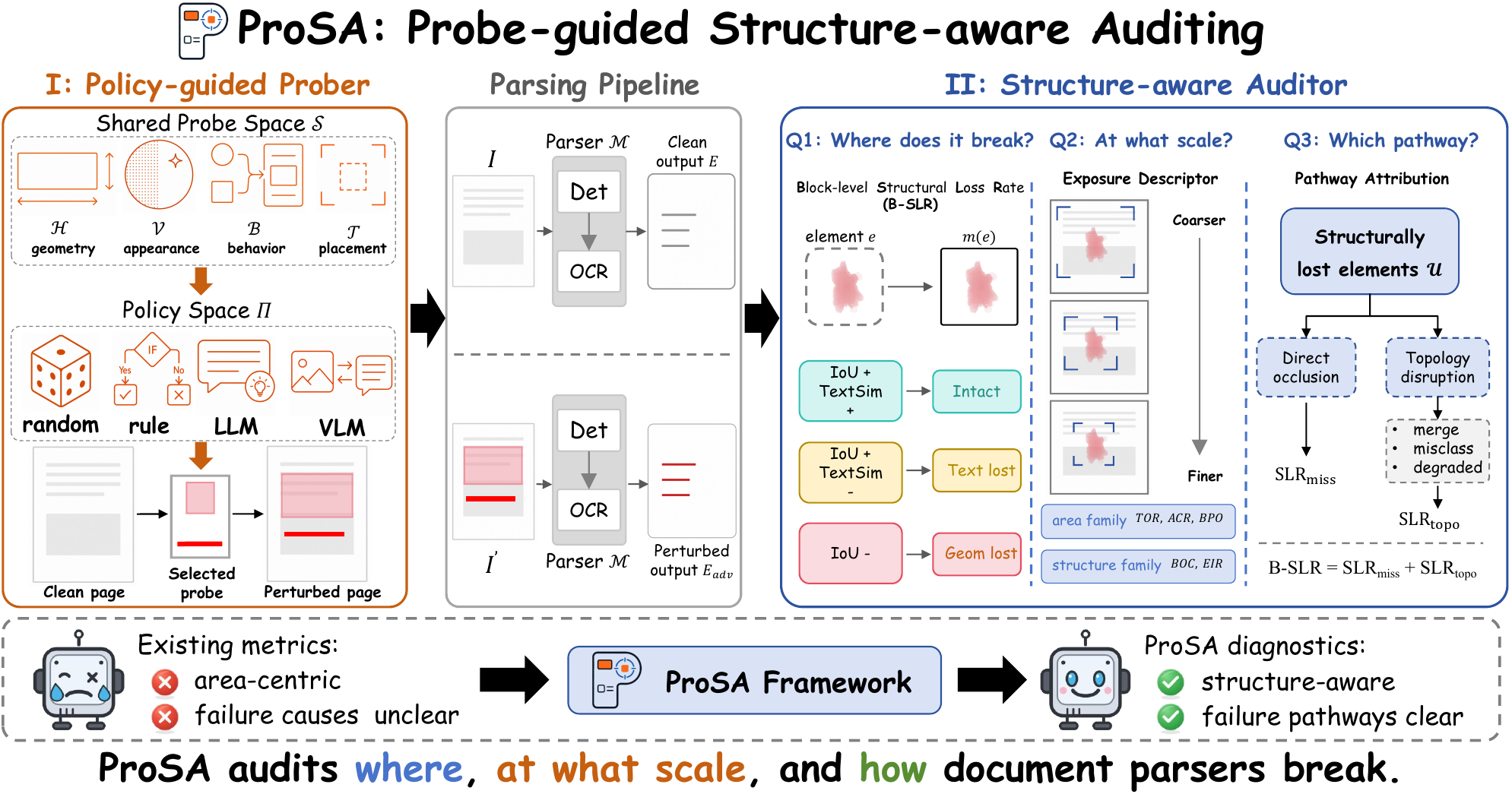}
  \caption{Overview of the proposed tripartite vulnerability auditing framework, linking controlled perturbation generation, policy-space probe selection, and diagnostic auditing.}
  \label{fig:framework_overview}
\end{figure*}

\section{Problem Setting}
\label{sec:problem_setting}

Let $\mathcal{M}$ denote a document layout analysis (DLA) system that maps a document image $I$ to a structured parsing output
\begin{equation}
E=\mathcal{M}(I)=\{e_i\}_{i=1}^{N},
\label{eq:dla_system}
\end{equation}
Each parsed element is a box--category--text tuple $e_i=(b_i,c_i,s_i)$: $b_i$ is its bounding box, $c_i\in\mathcal{C}_5$ its canonical label, and $s_i$ its recognized text. The five categories are text, title, table, figure, and equation.
This grounded box--category--text tuple defines the current ProSA audit contract: the parser must expose spatially grounded, text-bearing elements with categories that can be mapped to a shared schema, but ProSA does not require access to parser internals.

Given a probe configuration $P$, we obtain a counterfactual page and its parser output:
\begin{equation}
I'=\operatorname{Perturb}(I;P), \qquad
E_{\mathrm{adv}}=\mathcal{M}(I').
\label{eq:perturbation_setup}
\end{equation}

We formulate auditing as comparing the clean and perturbed parsing outputs:
\begin{equation}
(\Delta,\mathcal{R})
=
\operatorname{Audit}(E,E_{\mathrm{adv}};P,\mathcal{Y}),
\label{eq:audit_setting}
\end{equation}
where \(\mathcal{Y}\) denotes optional benchmark references when available.
The terminal component \(\Delta\) records externally visible degradation, including perturbation-induced OCR CER and detection-side \(\Delta\mathrm{mAP}\), while the structural component \(\mathcal{R}\) diagnoses where structural identity is lost, at what exposure granularity failure becomes predictive, and through which pathway it occurs.
In our experiments, CER is computed against the clean parser output, and \(\Delta\mathrm{mAP}\) is the per-image mAP@0.5 drop from clean to perturbed outputs against layout annotations.

This formulation extends conventional robustness evaluation beyond reporting \(\Delta\): by introducing \(\mathcal{R}\), the audit shifts from aggregate degradation measurement to structural perturbation diagnosis, without parser retraining or access to parser internals.

\section{Probe-guided Structure-aware Auditing}
\label{sec:framework}

% \subsection{Overview}
% \label{sec:framework_overview}

As illustrated in Figure~\ref{fig:framework_overview}, we propose \textbf{ProSA} (\textbf{Pro}be-guided \textbf{S}tructure-aware \textbf{A}uditing), a lightweight output-level framework for auditing structural vulnerability in DLA pipelines.
ProSA organizes the audit into two coupled components: a \emph{Policy-guided Prober}, which constructs controlled perturbations, and a \emph{Structure-aware Auditor}, which diagnoses the resulting clean--perturbed output divergence.
Formally, we instantiate ProSA as
\begin{equation}
\mathcal{F}=(\mathcal{S},\Pi,\mathcal{D}),
\label{eq:framework_overview}
\end{equation}
where $\mathcal{S}$ is a shared probe space, $\Pi$ is a policy space for selecting probes, and $\mathcal{D}$ is a diagnostic space for auditing the resulting parsing changes.
This decomposition separates \emph{what} perturbations can be applied, \emph{how} they are selected, and \emph{how} their effects are measured, enabling DLA parsers to be audited without modifying the parser itself.

\subsection{Component I: Policy-guided Prober}
\label{sec:controlled_perturbation}

The Policy-guided Prober defines the controlled perturbation side of ProSA.
It contains two coupled parts: a shared probe space $\mathcal{S}$ that defines admissible visual interventions, and a policy space $\Pi$ that selects probes under different targeting assumptions.
Because all policies draw from the same $\mathcal{S}$, differences in observed damage can be attributed to selection logic rather than search-space changes.

\paragraph{Shared probe space.}
Each probe configuration $P\in\mathcal{S}$ is represented as
\begin{equation}
P=\langle \mathcal{H},\mathcal{V},\mathcal{B},\mathcal{T}\rangle,
\label{eq:probe_tuple}
\end{equation}
where $\mathcal{H}$ denotes geometry, $\mathcal{V}$ visual appearance, $\mathcal{B}$ composition behavior, and $\mathcal{T}$ placement strategy.
This factorization keeps the perturbation space modular: geometry controls the spatial support, appearance controls the visible pattern, behavior controls how the probe is composed with the page, and placement controls where it interacts with layout structure.
Figure~\ref{fig:probe_demo} in Appendix~\ref{sec:probe_space} visualizes representative perturbation examples, while detailed probe definitions and parameter ranges are provided there.

\paragraph{Policy space.}
Given the shared probe space, a policy $\pi\in\Pi$ maps available document context to a selected probe:
\begin{equation}
\pi(\text{context})=\hat{P}\in\mathcal{S}.
\label{eq:policy_mapping}
\end{equation}
Different policies use different levels of context, ranging from random and rule-based selection to LLM- and VLM-guided targeting.
All policies emit the same probe schema, so their effects can be compared under a common action space.
After a policy selects $\hat{P}$, the perturbation operator applies it to the clean page to obtain $I'=\mathrm{Perturb}(I;\hat{P})$, and the same parser is rerun to produce $E_{\mathrm{adv}}=\mathcal{M}(I')$.
The resulting clean--perturbed pair is then passed to the Structure-aware Auditor.
Implementation details are deferred to Appendix~\ref{sec:policy_appendix}.

\subsection{Component II: Structure-aware Auditor}
\label{sec:auditor}

\paragraph{Diagnostic space.}
The Structure-aware Auditor instantiates the diagnostic space $\mathcal{D}$ by separating \emph{judging} from \emph{explaining}.
The terminal component $\Delta$ records externally visible degradation signals, while the structural component $\mathcal{R}$ diagnoses how degradation arises inside the parsing process:
\begin{equation}
\begin{aligned}
\Delta
&=\bigl(\Delta_{\mathrm{ocr}},\Delta_{\mathrm{det}}\bigr),\\
\mathcal{R}
&=\bigl(
\mathcal{R}_{\mathrm{fail}},
\mathcal{R}_{\mathrm{exposure}},
\mathcal{R}_{\mathrm{path}}
\bigr).
\end{aligned}
\label{eq:diagnostic_components}
\end{equation}
We instantiate \(\Delta_{\mathrm{ocr}}\) with OCR CER and \(\Delta_{\mathrm{det}}\) with
\(\Delta\mathrm{mAP}\), the per-image mAP@0.5 drop against layout annotations between the clean and perturbed outputs, following the metric definitions in Appendix~\ref{sec:terminal_metric_appendix}.
CER serves as the primary terminal signal, while $\Delta\mathrm{mAP}$ provides a supplementary detection-side check.
The three structural components correspond to the questions introduced in Section~\ref{sec:intro}: $\mathcal{R}_{\mathrm{fail}}$ measures whether structural identity is lost, $\mathcal{R}_{\mathrm{exposure}}$ describes at what granularity perturbation exposure becomes predictive of failure, and $\mathcal{R}_{\mathrm{path}}$ attributes failures to distinct propagation pathways.

\paragraph{Structural failure measurement.}
\label{sec:failure_measurement}

We instantiate $\mathcal{R}_{\mathrm{fail}}$ by checking whether each clean parsed element preserves a valid geometry-text correspondence after perturbation. For a clean element $e=(b,c,s)\in E$, we assign its candidate perturbed counterpart by best-IoU lookup:
\begin{equation}
m(e)=\arg\max_{e' \in E_{\mathrm{adv}}}\mathrm{IoU}(e,e'),
\label{eq:match}
\end{equation}
where $\mathrm{IoU}(e,e')$ denotes box overlap.

Geometric overlap alone is insufficient, because a perturbed prediction may still overlap the correct region while losing textual identity. We therefore define text consistency as
\begin{equation}
\mathrm{TextSim}(e,e')
=
\frac{
\left|\mathrm{LCS}\!\bigl(s(e),s(e')\bigr)\right|
}{
\max\!\bigl(|s(e)|,\;|s(e')|\bigr)
},
\label{eq:textsim}
\end{equation}
where $\mathrm{LCS}$ denotes character-level longest common subsequence. We write $e\sim e'$ when both geometry and text are preserved:
\begin{equation}
\begin{aligned}
e\sim e'
:=\;&
\bigl(\mathrm{IoU}(e,e')\ge\tau_{\mathrm{iou}}\bigr)\\
&\wedge
\bigl(\mathrm{TextSim}(e,e')\ge\tau_{\mathrm{text}}\bigr),
\end{aligned}
\label{eq:joint_align}
\end{equation}
with fixed diagnostic gates $\tau_{\mathrm{iou}}=0.1$ and $\tau_{\mathrm{text}}=0.5$, where IoU admits displaced counterparts and TextSim checks textual identity; see Appendix~\ref{sec:opconstants}.

Using this criterion, \textbf{B}lock-level \textbf{S}tructural \textbf{L}oss \textbf{R}ate (\textbf{B-SLR}) is defined as
\begin{equation}
\mathcal{R}_{\mathrm{fail}}
:=
\mathrm{B\mbox{-}SLR}
=
\frac{
\left|
\{e\in E:\neg(e\sim m(e))\}
\right|
}{|E|}.
\label{eq:bslr}
\end{equation}
Thus, B-SLR counts an element as structurally lost not only when its box disappears, but also when a nearby prediction remains while its textual identity is no longer preserved. Because both $E$ and $E_{\mathrm{adv}}$ are parser outputs, B-SLR does not require manual element annotations and can be computed on unlabeled documents using the clean parse as the structural reference.
Very short blocks form a local TextSim sensitivity corner case, but disabling the text gate for short blocks does not materially change configuration-level B-SLR/CER alignment; the full analysis is reported in Appendix~\ref{sec:shortblock_sensitivity}.

\paragraph{Granularity-aware exposure descriptors.}
\label{sec:damage_descriptors}

To explain what was exposed, we compare raw footprint with progressively more structure-aware descriptors. Total Occlusion Ratio (TOR) measures the fraction of page pixels affected; Area Coverage Ratio (ACR) measures affected area within annotated layout regions; Boundary Pixel Overlap (BPO) captures interaction with layout boundaries; Block Overlap Count ratio (BOC) and Element Interference Ratio (EIR) measure how many annotated or parser-visible elements are touched. TOR and EIR are output-derived, whereas ACR, BPO, and BOC require benchmark annotations. Formal definitions are provided in Appendix~\ref{sec:diag_appendix}.

\paragraph{Failure pathway attribution.}
\label{sec:pathway_attribution}

We partition structurally failed elements by whether the perturbation substantially covers the clean element. Failures with at least 30\% direct box coverage form the direct-occlusion pathway, SLR$_{\mathrm{miss}}$; the remaining failures form the topology-dominant pathway, SLR$_{\mathrm{topo}}$. These two disjoint pathways partition B-SLR, with formal definitions deferred to Appendix~\ref{sec:bslr_pseudocode}.
The topology-dominant pathway is further decomposed into merge, misclassification, and degradation subtypes; their formal definitions and assignment priority are given in Appendix~\ref{sec:bslr_pseudocode}.

\subsection{Validation principle}
\label{sec:validation_principle}

We validate each diagnostic by its configuration-level alignment and rank consistency with externally observable CER and detection degradation; the six-layer verification design is summarized in Appendix~\ref{sec:stats_appendix}.

\section{Experiments}

\subsection{Experimental Setup}
\label{sec:experimental_setup}

We instantiate $\mathcal{F}=(\mathcal{S},\Pi,\mathcal{D})$ on a shared pool of 1{,}000 validation pages from PubLayNet~\citep{publaynet2019} and DocLayNet~\citep{doclaynet2022}, after filtering pages with fewer than five original spans; the simple/medium/complex split is 14.5\%/46.4\%/39.1\%.

\paragraph{Audited pipelines.}
\textbf{Primary evaluation.} We conduct the full audit on \textbf{MinerU v2.7.6}~\citep{mineru2024}, based on DocLayout-YOLO~\citep{doclayoutyolo2024} with separate recognition modules, and \textbf{PP-StructureV3}~\citep{paddleocr2025}, a cascaded PaddleOCR pipeline using RT-DETR-H and PaddleOCR 3.x. Both expose grounded box--category--text elements and are evaluated over the shared 1{,}000-page pool through the complete Phase~1 and Phase~2 protocols using the same B-SLR matching thresholds ($\tau_{\mathrm{iou}}=0.1$, $\tau_{\mathrm{text}}=0.5$).

\paragraph{Grounded-VLM scope extension.}
We additionally audit \textbf{PaddleOCR-VL-1.6}~\citep{paddleocrvl2026} on a frozen stratified subset of 80 pages containing 221 QA pairs. Its \texttt{block\_bbox}, \texttt{block\_label}, and \texttt{block\_content} fields are mapped to the same audit contract without accessing model internals. This supplementary matched-footprint experiment tests interface compatibility; it is not part of the 1{,}000-page Phase~1/Phase~2 evaluation and does not support comparisons of parser robustness. Annotations are used only for sampling, detection metrics, and annotation-dependent exposure descriptors.

\paragraph{Auditing protocols.}
The controlled probe catalog (Table~\ref{tab:probe_catalog}) comprises real-world-inspired, factor-isolating counterfactual interventions over geometry, appearance, composition, and placement. These probes are designed to isolate affected area from parser-visible structural interaction; they do not reproduce complete physical capture processes or sample natural artifact distributions. We separately test whether the resulting B-SLR--CER relationship transfers to four standard corruption families (Appendix~\ref{sec:standard_corruption_transfer}).
For the two primary parsers, we use two complementary protocols. \textbf{Phase~1} performs fixed-configuration auditing by traversing 29 probe configurations in $\mathcal{S}$, yielding 29{,}000 records per pipeline. \textbf{Phase~2} performs policy-driven auditing by comparing five policy families---random, rule-based, LLM-biased, LLM-neutral, and VLM policy---over the same page pool and shared probe space. Configuration identifiers and policy details are provided in Appendices~\ref{sec:config_matrix} and~\ref{sec:policy_appendix}.

\paragraph{Statistical protocol.}
We report configuration-level analyses in the main text. Appendix~\ref{sec:stats_appendix} summarizes complementary verification layers; Appendices~\ref{sec:random_sweep_results} and~\ref{sec:dose_response_appendix} present randomized-sweep and dose-response results, respectively.

% \begin{table}[!htbp]
% \centering
% \small
% \resizebox{\columnwidth}{!}{%
% \begin{tabular}{lcc}
% \toprule
% Setting & MinerU & PP-V3 \\
% \midrule
% Input images / valid pages & 1{,}000 & 1{,}000 \\
% Shared evaluation pool & \multicolumn{2}{c}{1{,}000} \\
% Complexity (Simple / Medium / Complex) & \multicolumn{2}{c}{14.5\% / 46.4\% / 39.1\%} \\
% Phase 1 configs / Phase 2 policies & \multicolumn{2}{c}{29 / 5} \\
% Matching thresholds & \multicolumn{2}{c}{$\tau_{\mathrm{iou}}\!=\!0.1,\ \tau_{\mathrm{text}}\!=\!0.5$} \\
% \bottomrule
% \end{tabular}%
% }
% \caption{Dataset and protocol summary.}
% \vspace{-3mm}
% \label{tab:setup}
% \end{table}

\subsection{Results and Analysis}
\label{sec:main_results}

We first establish the mismatch between footprint and structural damage, then examine its downstream consequences and practical value for budgeted triage. We next test whether the audit contract transfers to a grounded VLM parser, before analyzing failure channels, exposure pathways, probe sensitivity, and targeting policies.

\Needspace{25\baselineskip}
\subsubsection{Footprint Bias Quantification}
\label{sec:footprint_bias}

\noindent\begin{minipage}{\columnwidth}
\centering
\includegraphics[width=\linewidth]{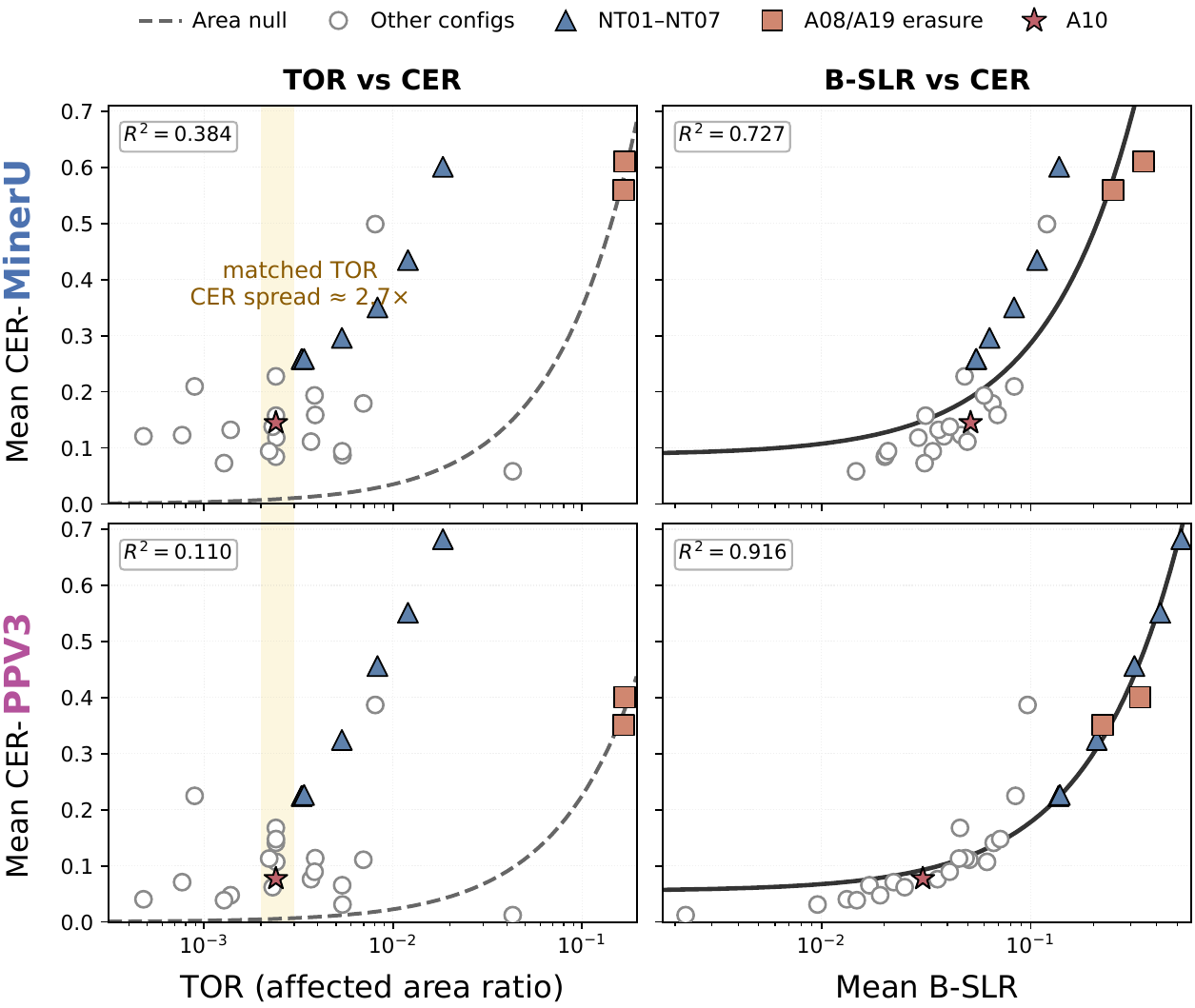}
\captionof{figure}{Configuration-level means ($n{=}29$) show that Block-level Structural Loss Rate (B-SLR) aligns with perturbation-induced Character Error Rate (CER) substantially better than affected area in both primary parsers; higher values indicate greater OCR instability and structural loss.}
\label{fig:footprint_bias}
\end{minipage}
\par\vspace{6pt}

Figure~\ref{fig:footprint_bias} tests whether affected area is a reliable severity proxy for perturbation-induced OCR instability.
It is not: TOR explains CER only weakly on MinerU ($R^2{=}0.384$) and almost not at all on PP-StructureV3 ($R^2{=}0.110$).
Even within the matched-TOR region, configurations with comparable footprint exhibit a CER spread of roughly $2.7\times$, showing that footprint alone cannot determine whether a perturbation is structurally harmful.

By contrast, realized structural loss measured by B-SLR is much more aligned with the terminal OCR signal, with $R^2{=}0.727$ on MinerU and $0.916$ on PP-StructureV3.
Configurations that appear anomalous under a footprint-only view return to the fitted trend once structural loss is considered.
Thus, the failure signal is not explained by how much area is perturbed alone, but by whether the perturbation breaks parser-visible structure.
B-SLR remains better aligned with CER than TOR under all nine tested $(\tau_{\mathrm{iou}},\tau_{\mathrm{text}})$ combinations; the complete sensitivity analysis is reported in Appendix~\ref{sec:threshold_sensitivity}.
Beyond the controlled-probe catalog, B-SLR remains strongly aligned with CER across 12 Gaussian-noise, Gaussian-blur, JPEG-compression, and brightness-shift configurations (Appendix~\ref{sec:standard_corruption_transfer}).

% Figure~\ref{fig:footprint_bias} directly tests whether affected area is a reliable predictor of OCR degradation, and shows that it is not.
% In the left column, TOR explains CER only weakly on MinerU ($R^2{=}0.384$) and almost not at all on PP-StructureV3 ($R^2{=}0.110$).
% Even within the matched-TOR region, configurations with comparable footprint still exhibit a CER spread of roughly $2.7\times$, and a footprint-only fit systematically underestimates many structurally targeted probes.

% In contrast, replacing TOR with B-SLR yields a much tighter relation to CER: $R^2{=}0.727$ on MinerU and $0.916$ on PP-StructureV3. Configurations that appear anomalously damaging under the area-based view return to the fitted trend once structural failure is used as the predictor. Thus, OCR degradation tracks structural damage more faithfully than raw footprint.

\subsubsection{Downstream Impact}
\label{sec:downstream}
\begin{table}[!htbp]
\centering
\scriptsize
\setlength{\tabcolsep}{3.0pt}
\renewcommand{\arraystretch}{0.92}
\resizebox{\columnwidth}{!}{
\begin{tabular}{@{}llrrrrrr@{}}
\toprule
Parser & Cond. & TOR & EM$\uparrow$ & Drop$\downarrow$ & Miss$\downarrow$ & BM25 A@5$\uparrow$ & Dense A@5$\uparrow$ \\
\midrule
\multirow{4}{*}{MinerU}
& Clean & 0.00  & 63.5 & --   & 0.0  & 94.6 & 93.6 \\
& AM    & 0.10  & 63.1 & 0.4  & 1.1  & 93.8 & 92.8 \\
& Str.  & 0.11  & 56.2 & \textbf{7.3} & \textbf{13.7} & \textbf{81.0} & \textbf{80.4} \\
& LA    & 16.61 & 38.8 & 24.7 & 36.8 & 58.8 & 59.4 \\
\midrule
\multirow{4}{*}{PP-V3}
& Clean & 0.00  & 63.6 & --   & 0.0  & 94.7 & 95.1 \\
& AM    & 0.10  & 63.1 & 0.5  & 2.3  & 92.9 & 93.4 \\
& Str.  & 0.11  & 58.1 & \textbf{5.5} & \textbf{12.2} & \textbf{83.1} & \textbf{83.5} \\
& LA    & 16.61 & 38.4 & 25.2 & 39.6 & 57.9 & 58.7 \\
\bottomrule
\end{tabular}
}
\caption{\textbf{Downstream propagation.}
All values are percentages. TOR: Total Occlusion Ratio; EM: exact match; Miss: answer-missing rate; A@5: answer hit at top 5; AM/Str./LA: area-matched erasure/structural probe/large-area erasure. Bold marks the worse AM--Str. result.}
\label{tab:downstream}
\end{table}

Table~\ref{tab:downstream} tests whether footprint-matched structural damage propagates to downstream document use. 
Rather than introducing a separate document QA benchmark, we reuse the same 975 clean-derived QA pairs under four counterfactual parser outputs, so that only the perturbation condition changes.
The primary QA results use \texttt{deepseek-chat}; two additional answerers preserve the positive area-matched-minus-structural (AM$-$Str.) EM/F1 gaps (Appendix~\ref{sec:downstream_appendix}).
For retrieval, the corpus is restricted to chunks from the same page and uses the same block-aware chunking across all conditions; this isolates whether answer-bearing evidence remains available after parsing.
The key contrast controls for visual footprint: area-matched erasure (AM) and the structural probe (Str.) perturb nearly the same page area, with TOR around 0.10--0.11\%.

Under this matched footprint, AM remains close to clean, whereas Str. consistently damages answering and retrieval. Across both primary parsers, Str. raises EM drop to 5.5--7.3 points and Miss to 12--14\%, while lowering BM25/Dense A@5 from above 92\% to 80--84\%. LA degrades most but uses about 150$\times$ more area. Thus, downstream failure follows structural interaction rather than footprint alone: a small structure-targeted probe can disrupt answer-bearing evidence where an area-matched erasure barely affects the task. Answerer-independent BM25 and dense-retrieval results further corroborate this trend; QA construction, filtering, type distribution, and type-wise gaps are detailed in Appendix~\ref{sec:downstream_appendix}.

\begingroup
\setlength{\parskip}{0pt}
\subsubsection{Budgeted Triage for Downstream Risk}
\label{sec:budgeted_triage}

ProSA audits rather than repairs. Under limited QA or retrieval budgets, high-risk pages may be inspected, re-OCRed, or routed to a fallback parser. We test whether its signals prioritize downstream harm by using each signal to rank 349 structural-probe pages and selecting the top $k=35$.

\vspace{4pt}
\Needspace{15\baselineskip}
\noindent\begin{minipage}{\columnwidth}
\centering
\footnotesize
\setlength{\tabcolsep}{3.1pt}
\begin{tabular}{llrrrr}
\toprule
Parser & Outcome & Random & TOR & EIR & B-SLR \\
\midrule
\multirow{3}{*}{MinerU}
  & QA EM & .104 & .042 & .099 & \textbf{.352} \\
  & BM25 ER@5 & .100 & .042 & .126 & \textbf{.137} \\
  & Dense ER@5 & .100 & .047 & .126 & \textbf{.137} \\
\midrule
\multirow{3}{*}{PP-V3}
  & QA EM & .103 & .037 & .241 & \textbf{.648} \\
  & BM25 ER@5 & .101 & .023 & .165 & \textbf{.323} \\
  & Dense ER@5 & .100 & .000 & .165 & \textbf{.331} \\
\bottomrule
\end{tabular}
\captionof{table}{\textbf{Fixed-budget downstream-risk triage.} Fraction of the clean-to-perturbed gap recovered by selecting 35 of 349 pages (10\% budget); higher is better. For selected pages, all associated QA and retrieval evaluations use the clean parser output; unselected pages retain the structural-probe output. Random is averaged over 1{,}000 seeded selections. QA EM: question-answering Exact Match; ER@5: Evidence Recall at 5; TOR/EIR/B-SLR: Total Occlusion Ratio/Element Interference Ratio/Block-level Structural Loss Rate.}
\label{tab:budgeted_triage}
\end{minipage}
\par\vspace{6pt}

Table~\ref{tab:budgeted_triage} evaluates external QA and retrieval rather than a ProSA-derived target. For a ranking signal $s$, let $\mathcal{K}_s$ denote its 35 highest-ranked pages. We simulate restoration by replacing the structural-probe parser output with its clean counterpart for every page in $\mathcal{K}_s$, while retaining the structural-probe output for all remaining pages. All QA instances and retrieval chunks associated with a selected page are replaced together.

For a higher-is-better downstream metric $M$, $M_{\mathrm{clean}}$ is the aggregate score when every page uses its clean parser output, $M_{\mathrm{structural}}$ is the score when every page uses its structural-probe output, and $M_{\mathrm{restored}}(s)$ is the score under the hybrid clean/structural corpus induced by $\mathcal{K}_s$. We define
\begin{equation}
\operatorname{Recovery}(s;M)
=
\frac{M_{\mathrm{restored}}(s)-M_{\mathrm{structural}}}
{M_{\mathrm{clean}}-M_{\mathrm{structural}}}.
\label{eq:triage_recovery}
\end{equation}
\Needspace{4\baselineskip}
Thus, zero indicates no improvement over the fully perturbed condition, whereas one indicates recovery of the entire clean--structural performance gap. The Random baseline averages this quantity over 1{,}000 independently sampled 35-page subsets.

EIR outperforms the footprint baseline TOR on all six outcomes, with the clearest gains on PP-StructureV3 QA (.241 vs. .037) and retrieval (.165/.165 vs. .023/.000). On MinerU, EIR improves retrieval (.126/.126 vs. .042/.047), while QA remains comparable to Random (.099 vs. .104). B-SLR provides the strongest recovery, but because it requires paired parser outputs, it serves post-output auditing or regression testing rather than earlier exposure-based triage.

\makeatletter
\begingroup
\def\subsubsection{\@startsection{subsubsection}{3}{\z@}{-1.5ex}{0.5ex}{\normalsize\bfseries\raggedright}}
\subsubsection{Extension to a Grounded VLM Parser}
\label{sec:vlm_scope_result}

To test whether ProSA is tied to the two primary parsers or to their shared output interface, we additionally audit PaddleOCR-VL-1.6, a recent grounded VLM parser that exposes block boxes, labels, and text. On the frozen 80-page subset with 221 QA pairs, the matched-footprint comparison yields a B-SLR gap of $.056\,[.021,.100]$, whose 95\% confidence interval excludes zero. This result shows that the same box--category--text audit can detect greater structural damage from structure-targeted perturbations than from area-matched erasure in this grounded VLM pipeline.

The downstream estimates point in the same direction: the answer-missing, BM25, and dense-retrieval gaps are $.023$, $.014$, and $.009$, respectively. However, all three 95\% confidence intervals include zero. We therefore treat the VLM experiment as evidence that ProSA's structural audit can be instantiated on a grounded VLM parser satisfying the required output interface, while the downstream evidence on this subset remains inconclusive. We do not use these results to claim significant downstream replication or to rank robustness across parsers. Full estimates and the 5{,}000-resample paired-bootstrap protocol are reported in Appendix~\ref{sec:vlm_scope_appendix}.
\endgroup
\makeatother
\endgroup

\subsubsection{B-SLR Channels and Faithfulness}
\label{sec:bslr_result}

Returning to the parser-level diagnostic, Table~\ref{tab:fail_channel} decomposes composite B-SLR into an IoU-fail channel, where no valid geometric match survives, and a text-fail-only channel, where geometry is retained but textual identity is lost. Text-side failure is non-redundant and dominant in both primary pipelines: it exceeds the IoU channel on MinerU and more strongly on PP-V3, showing that many structural failures preserve coarse geometry while losing recognized content.

\begin{table}[t]
\centering
\small\raggedright\hyphenpenalty=10000\exhyphenpenalty=10000
\setlength{\tabcolsep}{3.2pt}
\renewcommand{\arraystretch}{1.1}
\captionsetup{position=top,skip=3pt}
\caption{Configuration-level Block-level Structural Loss Rate (B-SLR) channel decomposition ($n{=}29$): mean channel losses and cross-channel coupling ($\rho_s$ for Intersection-over-Union (IoU)--text rank correlation; $R^2$ columns are univariate Ordinary Least Squares (OLS) channel--Character Error Rate (CER) fits).}
\label{tab:fail_channel}
\begin{minipage}{0.48\linewidth}\centering
\textbf{(a) Channel partition}\\[2pt]
\begin{tabular}{@{}lccc@{}}
\toprule
 & B-SLR & IoU & text \\
\midrule
MinerU & .071 & .028 & \textbf{.043} \\
PP-V3  & .111 & .026 & \textbf{.085} \\
\bottomrule
\end{tabular}
\end{minipage}\hfill
\begin{minipage}{0.48\linewidth}\centering
\textbf{(b) Cross-channel coupling}\\[2pt]
\begin{tabular}{@{}lccc@{}}
\toprule
 & $\rho_{s}$ & $R^{2}_{\mathrm{IoU}}$ & $R^{2}_{\mathrm{text}}$ \\
\midrule
MinerU & .891 & \textbf{.927} & .893 \\
PP-V3  & .380 & .186 & \textbf{.900} \\
\bottomrule
\end{tabular}
\end{minipage}
\end{table}

The channel coupling differs sharply across architectures. MinerU shows coherent geometry--text variation ($\rho_s{=}.891$), whereas PP-V3 is driven mainly by the text channel ($R^2_{\mathrm{text}}{=}.900$ vs.\ $R^2_{\mathrm{IoU}}{=}.186$).

Thus, B-SLR is not merely a text-only score. It captures architecture-specific coupling between geometry and text failures while remaining aligned with OCR instability. The resulting $R^2$ values are $.727/.916$ for MinerU/PP-V3, with Spearman $\rho{=}.911/.973$.

\setcounter{table}{4}
\begin{table*}[!t]
\centering
\scriptsize
\setlength{\tabcolsep}{4pt}
\newcommand{\torcell}[1]{\cellcolor{red!8}{#1}}
\resizebox{\textwidth}{!}{%
\begin{tabular}{ll cccc ccc ccc cc}
\toprule
\multirow{2}{*}{Parser} & \multirow{2}{*}{Policy} &
\multicolumn{4}{c}{Exposure / Budget} &
\multicolumn{3}{c}{Damage} &
\multicolumn{3}{c}{Failure Pathway} &
\multicolumn{2}{c}{Efficiency} \\
\cmidrule(lr){3-6} \cmidrule(lr){7-9} \cmidrule(lr){10-12} \cmidrule(lr){13-14}
& & \torcell{TOR} & BPO & BOC & EIR
& CER$\uparrow$ & $\Delta$mAP$\uparrow$ & B-SLR$\uparrow$
& Miss & Topo & TopoShare
& Eff-B$\uparrow$ & Eff-C$\uparrow$ \\
\midrule
\multirow{5}{*}{MinerU}
 & Random      & \torcell{.035} & .050 & .331 & .357 & .226 & .043 & .088 & .018 & .070 & .82  & 10.9          & 24.5 \\
 & Rule-based  & \torcell{.005} & .015 & .292 & .352 & .245 & .048 & .077 & .000 & .077 & 1.00 & 23.8          & 61.4 \\
 & LLM-biased  & \torcell{.006} & .026 & .157 & .182 & .177 & .025 & .068 & .003 & .065 & .94  & 16.0          & 35.7 \\
 & LLM-neutral & \torcell{.008} & .011 & .125 & .145 & .209 & .034 & .058 & .003 & .055 & .91  & 12.2          & 40.2 \\
 & VLM policy  & \torcell{.003} & .007 & .099 & .124 & .176 & .014 & .056 & .000 & .056 & 1.00 & \textbf{25.3} & \textbf{73.3} \\
\midrule
\multirow{5}{*}{PP-V3}
 & Random      & \torcell{.035} & .050 & .331 & .312 & .227 & .026 & .077 & .013 & .064 & .80 & 9.1           & 22.1 \\
 & Rule-based  & \torcell{.005} & .015 & .292 & .312 & .230 & .037 & .077 & .001 & .076 & .98 & \textbf{17.6} & \textbf{33.3} \\
 & LLM-biased  & \torcell{.007} & .027 & .155 & .125 & .123 & .012 & .049 & .003 & .046 & .88 & 9.4           & 18.1 \\
 & LLM-neutral & \torcell{.008} & .012 & .127 & .129 & .119 & .024 & .048 & .004 & .044 & .83 & 10.7          & 20.5 \\
 & VLM policy  & \torcell{.003} & .007 & .104 & .092 & .081 & .005 & .034 & .000 & .033 & .98 & 15.9          & 30.5 \\
\bottomrule
\end{tabular}
}
\caption{\textbf{Phase~2 policy audit.}
Under the shared probe space and page pool, targeted policies reveal the expected efficiency pattern. TOR/BPO/BOC/EIR: Total Occlusion Ratio/Boundary Pixel Overlap/Block Overlap Count ratio/Element Interference Ratio; CER: Character Error Rate; $\Delta$mAP: mean Average Precision drop; B-SLR: Block-level Structural Loss Rate; LLM/VLM: Large Language Model/Vision-Language Model; Miss/Topo: direct-occlusion/topology-dominant loss; TopoShare: topology-dominant loss share; Eff-B/Eff-C: B-SLR/CER per unit TOR. Metric details are provided in Appendix~\ref{sec:policy_metric_shorthands}.
}
\label{tab:policy_main}
\end{table*}
\setcounter{table}{3}

\subsubsection{Exposure Descriptors Predict Failure Pathways}
\label{sec:exposure_result}

\begin{center}
\begin{minipage}{\columnwidth}
\centering
\small\raggedright\hyphenpenalty=10000\exhyphenpenalty=10000
\setlength{\tabcolsep}{4pt}
\begin{tabular}{l l ccc cc}
\toprule
 & & \multicolumn{3}{c}{$\mathcal{G}_{\text{area}}$} & \multicolumn{2}{c}{$\mathcal{G}_{\text{struct}}$} \\
\cmidrule(lr){3-5}\cmidrule(lr){6-7}
Pipeline & Pathway & TOR & ACR & BPO & BOC & EIR \\
\midrule
\multirow{2}{*}{MinerU}
 & SLR$_{\mathrm{miss}}$ & .927 & .907 & \textbf{.964} & .071 & .044 \\
 & SLR$_{\mathrm{topo}}$ & .004 & .005 & .003 & .610 & \textbf{.623} \\
\addlinespace
\multirow{2}{*}{PP-V3}
 & SLR$_{\mathrm{miss}}$ & .916 & .895 & \textbf{.960} & .072 & .073 \\
 & SLR$_{\mathrm{topo}}$ & .009 & .008 & .008 & \textbf{.708} & .702 \\
\bottomrule
\end{tabular}
\captionof{table}{Configuration-level $R^2$ ($n{=}29$) from univariate Ordinary Least Squares (OLS) fits between exposure descriptors and pathway loss. TOR/ACR/BPO: Total Occlusion Ratio/Area Coverage Ratio/Boundary Pixel Overlap; BOC/EIR: Block Overlap Count ratio/Element Interference Ratio; SLR$_{\mathrm{miss}}$/SLR$_{\mathrm{topo}}$: direct-occlusion/topology-dominant Structural Loss Rate.}
\label{tab:exposure_pathway}
\end{minipage}
\end{center}
% The Phase~2 table is declared earlier so it can float to the top of the next page.
% Its number is reserved as Table~\ref{tab:policy_main}; restore the counter after Table~\ref{tab:exposure_pathway}.
\setcounter{table}{5}
\hspace*{\parindent}We next ask which exposure granularity predicts each failure pathway. 
Table~\ref{tab:exposure_pathway} regresses the two pathway losses, SLR$_{\mathrm{miss}}$ and SLR$_{\mathrm{topo}}$, on the five diagnostic descriptors, grouped into the area family $\mathcal{G}_{\text{area}}=(\text{TOR},\text{ACR},\text{BPO})$ and the structure family $\mathcal{G}_{\text{struct}}=(\text{BOC},\text{EIR})$.

\indent The specialization is consistent across both pipelines. 
For SLR$_{\mathrm{miss}}$, BPO is the strongest predictor ($R^2{=}0.964/0.960$ on MinerU/PP-V3), with TOR and ACR also high but weaker. 
For SLR$_{\mathrm{topo}}$, the pattern reverses: area-family descriptors drop near zero, whereas BOC/EIR dominate ($R^2{=}0.610/0.623$ on MinerU and $0.708/0.702$ on PP-V3). 
Thus, the two descriptor families are not interchangeable severity measures: boundary-sensitive area exposure predicts direct occlusion, while element-level structural interference predicts topology-dominant loss.
A complementary fixed-budget analysis over 42{,}000 fixed-plus-sweep records per parser recovers the same descriptor specialization at Precision@10\% (Appendix~\ref{sec:extended_tables}).

\indent This addresses Q$_2$: footprint becomes informative only after being refined by structural contact.
As exposure descriptors computed before terminal scoring, BPO and EIR indicate different risks: boundary overlap signals direct occlusion, while element interference signals topology-dominant disruption. 
The NT sweep in Section~\ref{sec:phase1_result} exemplifies this regime, increasing structural interference without a matched increase in TOR.
A ranking-consistency check against CER yields the same family-level ordering, indicating that the specialization is not an artifact of B-SLR alone.

\subsubsection{Sensitivity Across Probe Families}
\label{sec:phase1_result}

\begin{figure}[!htbp]
  \centering
  \includegraphics[width=\columnwidth]{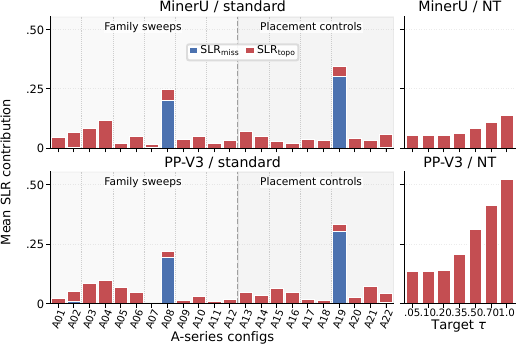}
  \caption{\textbf{Phase~1 pathway decomposition.}
  Bars decompose Block-level Structural Loss Rate (B-SLR) into direct-occlusion SLR$_{\mathrm{miss}}$ and topology-dominant SLR$_{\mathrm{topo}}$; higher bars indicate greater structural loss, and configuration identifiers are decoded in Appendix~\ref{sec:config_matrix}.}
  \label{fig:phase1_pathway_main}
\end{figure}

\begingroup
\fussy
Figure~\ref{fig:phase1_pathway_main} reports the Phase~1 fixed-configuration audit, where the targeting rule is fixed and variation mainly reflects probe-level effects. Damage is highly concentrated: the largest A-series spikes appear at A08 and A19 for both pipelines, indicating that severe erasure-style probes are disproportionately destructive, while most other configurations remain much weaker. The dominant pathway also varies by configuration: extreme erasure is largely miss-driven, whereas many non-extreme configurations are dominated by $\mathrm{SLR}_{\mathrm{topo}}$, indicating topology-level disruption rather than direct element removal.

The NT sweep further reveals parser-dependent sensitivity to sustained structural interference. MinerU increases gradually as the target $\tau$ increases, whereas PP-StructureV3 shows a steeper rise in topology-dominant loss at higher target levels. Overall, Phase~1 shows that the probe space contains qualitatively different failure regimes: severe erasure produces miss-heavy spikes, while structure-targeted interference increasingly induces topology-dominant failure.
\endgroup

\subsubsection{Targeting Efficiency Across Policies}
\label{sec:phase2_result}

\hspace*{\parindent}Table~\ref{tab:policy_main} reports the Phase~2 policy audit, where all policies share the same probe space and page pool, so differences reflect targeting logic rather than attack strength.
Targeted policies consistently trade footprint for structural contact: Rule-based matches Random on PP-V3 B-SLR with about one seventh of the TOR, and the VLM policy/Rule-based policy achieve the highest per-footprint efficiencies on MinerU/PP-V3.
Their high TopoShare and near-zero Miss indicate that this efficiency mainly comes from topology-sensitive disruption rather than direct physical occlusion.
Thus, under a shared $\mathcal{S}$, vulnerability depends not only on how much area a policy perturbs, but on whether it places probes where layout structure is sensitive; Appendix~\ref{sec:policy_rank_consistency} shows the same rank-consistency trend.

\section{Conclusion}
We identify \emph{Footprint Bias} and introduce \textbf{ProSA}, an output-level framework for auditing parser-visible structural loss.
Across the two primary parsers, B-SLR tracks OCR instability more faithfully than affected area, and structure-targeted probes cause disproportionate downstream degradation.
The fixed-budget simulation further shows that ProSA signals can support downstream-risk triage, while the supplementary PaddleOCR-VL-1.6 audit provides initial compatibility evidence for a grounded VLM parser.
These findings support structure-aware auditing without claiming automated repair, universal VLM coverage, or comparative parser robustness.

% Required by ACL; excluded from the content page limit.
\section*{Limitations}
ProSA is a diagnostic audit, not universal robustness certification.
First, the full audit covers two DLA pipelines, and the grounded-VLM extension uses a smaller frozen subset; the results show within-parser diagnostic behavior and interface compatibility, not broad cross-parser rankings.
Second, B-SLR uses the clean parser output as its reference, so it measures induced instability but misses errors already in, or shared with, the clean output.
Third, ProSA requires spatially grounded box--category--text elements; serialization-only or weakly structured systems need an alignment adapter. The reported pathways are output-level attributions, not internal causal claims.

% Optional under ACL; excluded from the content page limit.
\section*{Ethical Considerations}
This work studies vulnerability auditing for defensive robustness evaluation. We report attack-like procedures solely to improve safety testing and mitigation in document intelligence systems.

% Must immediately precede the references; excluded from the content page limit.
\section*{Acknowledgments}
{\setlength{\baselineskip}{12.55pt}
This work was supported by the National Natural Science Foundation of China (NSFC; Grant 62276283), the China Meteorological Administration's Science and Technology Project (CMAJBGS202517), the Guangdong--Hong Kong--Macao Greater Bay Area Meteorological Technology Collaborative Research Project (GHMA2024Z04), the Fundamental Research Funds for the Central Universities, Sun Yat-sen University (23hytd006 and 23hytd006-2), the Guangdong Provincial High-Level Young Talent Program (RL2024-151-2-11), the Key Development Project of the Artificial Intelligence Institute, Sun Yat-sen University, and the Major Key Project of PCL (PCL2025A17). We thank the anonymous reviewers and area chairs for their constructive feedback. AI assistants were used for language polishing, editing assistance, and code debugging; all outputs were independently checked and verified by the authors.
\par}
\vspace{-6.200pt}

\bibliography{reference}

\appendix
\raggedbottom
\makeatletter
\renewcommand\section{\@startsection {section}{1}{\z@}%
  {-1.6ex plus -.4ex minus -.2ex}%
  {0.7ex plus .2ex}%
  {\normalfont\large\bfseries}}
\renewcommand\subsection{\@startsection{subsection}{2}{\z@}%
  {-1.1ex plus -.3ex minus -.2ex}%
  {0.45ex plus .1ex}%
  {\normalfont\normalsize\bfseries}}
\makeatother

%% ================================================================
%%  Appendix A — Methodological Details
%% ================================================================

\section{Implementation Details of Probe and Policy Spaces}
\label{sec:appendix_probe_policy}

\begin{table*}[tbp]
\centering
\small\raggedright\hyphenpenalty=10000\exhyphenpenalty=10000
\setlength{\tabcolsep}{5pt}
\begin{tabularx}{\textwidth}{@{}p{0.18\textwidth}p{0.09\textwidth}p{0.09\textwidth}Yp{0.15\textwidth}@{}}
\toprule
Probe & Geom. & Behavior & Key parameters & Placement $\mathcal{T}$ \\
\midrule
P1 (horiz.\ crease) & line & inject & $w\!\in\![1,10],\; l_r\!\in\![0.5,1.0]$ & \texttt{a}/\texttt{c}/\texttt{r} \\
P2 (vert.\ crease) & line & inject & $w\!\in\![1,10],\; l_r\!\in\![0.5,1.0]$ & \texttt{a}/\texttt{c}/\texttt{r} \\
P3 (circular overlay) & disk & blend & $r\!\in\![30,90],\; \alpha\!\in\![0.2,1.0]$ & \texttt{a}/\texttt{c}/\texttt{r} \\
P4 (rect.\ modification) & rect & erase & $a_{\mathrm{area}}\!\in\![3\%,25\%],\; \beta\!\in\![0.2,1.0]$ & \texttt{c}/\texttt{b} \\
P5 (thin horiz.\ line) & line & inject & $w\!\in\![1,5],\; l_r\!\in\![0.2,0.8]$ & \texttt{b}/\texttt{c}/\texttt{r} \\
P6 (gradient band) & line+grad & blend & $\alpha\!\in\![0.05,0.4],\; w\!\in\![2,10]$ & \texttt{a} \\
P7 (dot cluster) & points & inject & $n\!\in\![10,100],\; r\!\in\![1,4],\; \sigma\!\in\![10,50]$ & \texttt{r} \\
P8 (irregular patch) & blob & blend & $r_b\!\in\![30,80],\; \kappa\!\in\![0.1,0.5],\; \alpha\!\in\![0.3,0.7]$ & \texttt{a} \\
P9 (diag.\ crease) & line & inject & $\theta\!\in\![20^\circ,70^\circ],\; w\!\in\![1,6]$ & \texttt{a} \\
\bottomrule
\end{tabularx}
\caption{Probe catalog and parameter ranges. Placement codes \texttt{a}/\texttt{c}/\texttt{r}/\texttt{b} denote anchor/content/random/bridge.}
\label{tab:probe_catalog}
\end{table*}

\subsection{Probe Space and Probe Catalog}
\label{sec:probe_space}

Each probe configuration is specified by four dimensions:
\begin{equation}
P=\langle \mathcal{H},\mathcal{V},\mathcal{B},\mathcal{T}\rangle,
\end{equation}
where $\mathcal{H}$ denotes geometry, $\mathcal{V}$ visual appearance, $\mathcal{B}$ composition behavior, and $\mathcal{T}$ placement strategy.
This factorization keeps the perturbation space modular: probe geometry, visual appearance, composition behavior, and placement can be controlled or varied while all policies draw from the same admissible space.

\begin{figure*}[tbp]
\centering
\includegraphics[width=\textwidth]{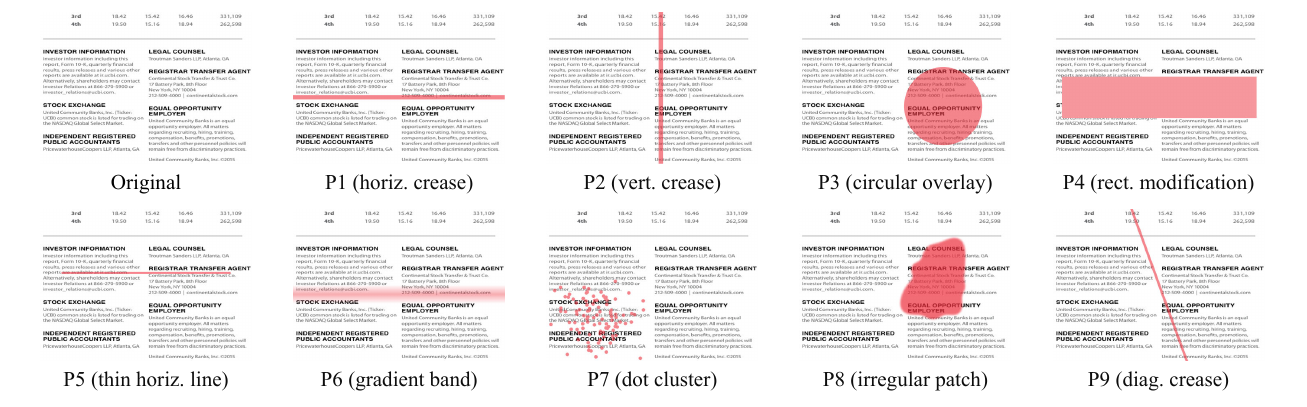}
\caption{Representative visual examples of the probe families used in the controlled perturbation space.}
\label{fig:probe_demo}
\end{figure*}

\begingroup
\fussy
Table~\ref{tab:probe_catalog} summarizes the probe used in our audit.
The catalog covers line-like artifacts, area erasures, local overlays, point clusters, and irregular patches, so that the shared probe space contains both footprint-heavy and structure-sensitive perturbations.
Figure~\ref{fig:probe_demo} visualizes representative examples of these probe families.
\par
\endgroup

\subsection{Probe Generation and Placement}
\label{sec:probe_generation}

This subsection specifies how the probe catalog in Table~\ref{tab:probe_catalog} is rendered on a document page. 
We denote geometry masks by $\mathcal{H}_{(\cdot)}$ to distinguish probe geometry from the diagnostic exposure families in the main text. 
The geometry primitives are implemented as binary masks:
\begin{alignat}{2}
&\mathcal{H}_{\mathrm{line}}(\theta,l,w) &&= \{(x,y):\notag\\
&                                      &&\phantom{={}} |x\sin\theta-y\cos\theta|<\tfrac{w}{2},\notag\\
&                                      &&\phantom{={}} 0\le x\cos\theta+y\sin\theta\le l\},\\[3pt]
&\mathcal{H}_{\mathrm{disk}}(r) &&= \{(x,y):x^2+y^2\le r^2\},\\[3pt]
&\mathcal{H}_{\mathrm{rect}}(w_r,h_r) &&= \{(x,y): |x|\le w_r/2,\notag\\
&                                      &&\phantom{={}} |y|\le h_r/2\},\\[3pt]
&\mathcal{H}_{\mathrm{blob}}(r_b,\kappa) &&= \{(r,\phi): r\le r_b\bigl(1+\kappa\cdot\notag\\
&                                      &&\phantom{={}} \mathrm{Perlin}(\phi)\bigr)\},\\[3pt]
&\mathcal{H}_{\mathrm{point}}(n,r,\sigma) &&= \bigcup_{i=1}^{n}\mathcal{H}_{\mathrm{disk}}(r;c_i),\notag\\
&                                      &&\phantom{={}} c_i\sim\mathcal{N}(\mu,\sigma^2I).
\end{alignat}

Appearance families include solid color, gradient, ring, and procedural texture.
All probe actions are unified by alpha blending:
\begin{equation}
I'(x,y)=\bigl(1-\alpha(x,y)\bigr)I(x,y)+\alpha(x,y)q(x,y),
\end{equation}
where $q(x,y)$ is the probe texture or background target.
The three composition behaviors in Table~\ref{tab:probe_catalog} are instantiated as \textit{inject} ($\alpha=1$), \textit{blend} ($\alpha\in(0,1)$), and \textit{erase} (blending toward the local background color).

Placement strategies are defined over content and boundary masks derived from the clean-layout boxes.
Let $\{b_i\}$ denote clean-layout boxes and $\partial b_i$ their boundaries.
The anchor region is
\begin{equation}
M_{\mathrm{anchor}}=\operatorname{Dil}\!\left(\bigcup_i\partial b_i,\delta\right)\setminus M_{\mathrm{content}}^{\mathrm{erode}(\delta)},
\end{equation}
where $\delta=5$\,px (Appendix~\ref{sec:opconstants}).
We use four placement strategies: \textbf{anchor} (\texttt{a}, on $M_{\mathrm{anchor}}$), \textbf{content} (\texttt{c}, inside $M_{\mathrm{content}}$), \textbf{random} (\texttt{r}, uniformly over the page), and \textbf{bridge} (\texttt{b}, at the midpoint between two neighboring blocks). 
Across experiments, one to three probes are placed per page under a global area budget.

\subsection{Policy Families}
\label{sec:policy_appendix}

The policy space $\Pi$ maps page context to a probe configuration in the shared probe space of Table~\ref{tab:probe_catalog}. 
Formally, each policy is written as
\begin{equation}
\pi(\text{context})
=
\hat{P}
=
\langle
\hat{\mathcal{H}},
\hat{\mathcal{V}},
\hat{\mathcal{B}},
\hat{\mathcal{T}}
\rangle,
\end{equation}
where the context features are extracted from the current page, and $\hat{P}$ follows the same geometry--appearance--behavior--placement schema as Section~\ref{sec:probe_space}.

We use five non-iterative policy families under the same probe space:
\begin{itemize}
    \item \textbf{Random} ($\pi_{\mathrm{rand}}$): uniformly samples the probe type, parameters, and placement strategy.
    \item \textbf{Rule-based} ($\pi_{\mathrm{rule}}$): selects probes using boundary density and inter-block gap heuristics.
    \item \textbf{LLM-biased} ($\pi_{\mathrm{LLM}}^{\mathrm{biased}}$): uses a language-guided prompt with explicit structural hints.
    \item \textbf{LLM-neutral} ($\pi_{\mathrm{LLM}}^{\mathrm{neutral}}$): uses a language-guided prompt without explicit structural hints.
    \item \textbf{VLM strategy-only} ($\pi_{\mathrm{VLM}}$): predicts probe type and placement strategy from the page image without direct coordinate output.
\end{itemize}
All policies emit the same schema---probe type, parameters, and placement strategy---so differences in observed damage can be attributed to targeting logic rather than search-space mismatch.

\subsection{Controlled Prompt Design}
\label{sec:prompt_control}

The biased and neutral LLM prompts test whether structural targeting arises from explicit hints or page context. 
They differ in role, objective, strategy labels, context rendering, and hint strength. 
A boundary preference under the neutral prompt therefore cannot be attributed solely to explicit structural hints.

\section{Diagnostic Metrics, Matching, and Terminal Evaluation}
\label{sec:appendix_diagnostics}

\subsection{Exposure Descriptor Definitions}
\label{sec:diag_appendix}

This subsection gives the operational definitions of the five exposure descriptors used in Section~\ref{sec:damage_descriptors}. 
Let $A(P)$ denote the perturbation mask induced by probe configuration $P$, and let $(W,H)$ be the page size. 
For annotated layout elements $L$, we define the annotated layout support and boundary support as
\begin{flalign}
&\Omega_L = \bigcup_{\ell\in L} b(\ell),\notag\\
&M_{\partial L} = \operatorname{Dil}\!\left(\cup_{\ell\in L}\partial b(\ell),\delta\right)\setminus \Omega_L^{\mathrm{erode}(\delta)} .&&
\end{flalign}
The five descriptors are computed as
\begin{flalign}
&\mathrm{TOR}(P)=\frac{|A(P)|}{W\times H},\notag\\
&\mathrm{ACR}(P,L)=\frac{|A(P)\cap \Omega_L|}{|\Omega_L|},\notag\\
&\mathrm{BPO}(P,L)=\frac{|A(P)\cap M_{\partial L}|}{|M_{\partial L}|},\notag\\
&\mathrm{BOC}(P,L)=\frac{|\{\ell\in L: b(\ell)\cap A(P)\neq\emptyset\}|}{|L|},\notag\\
&\mathrm{EIR}(P,E)=\frac{|\{e\in E: b(e)\cap A(P)\neq\emptyset\}|}{|E|}.&&
\label{eq:descriptor_ops}
\end{flalign}
TOR and EIR are output-derived, whereas ACR, BPO, and BOC are annotation-dependent benchmark diagnostics. 
For BOC and EIR, the intersection predicate is satisfied when at least one pixel of the bounding box overlaps the probe mask (\texttt{overlap\_px=1}).

\subsection{Output Matching and TextSim}
\label{sec:matching_appendix}

Each clean element $e\in E$ is assigned a candidate counterpart in the perturbed output by maximum-IoU lookup:
\begin{equation}
m(e)=\arg\max_{e'\in E_{\mathrm{adv}}}\mathrm{IoU}(e,e').
\end{equation}
Ties are broken by adversarial-element index, and no one-to-one Hungarian assignment is enforced. 
This allows multiple clean elements to map to the same perturbed element, which is necessary for detecting merge-type topology failures.

A match is accepted only when both the geometry and text gates are satisfied:
\begin{equation}
\begin{aligned}
\mathrm{IoU}(e,m(e))&\ge\tau_{\mathrm{iou}},\\
\mathrm{TextSim}(e,m(e))&\ge\tau_{\mathrm{text}}.
\end{aligned}
\end{equation}
We use $\tau_{\mathrm{iou}}=0.1$ and $\tau_{\mathrm{text}}=0.5$ in all experiments; their rationale and sensitivity are discussed in Appendix~\ref{sec:threshold_sensitivity}.

Before computing text consistency, both strings are stripped of leading/trailing whitespace and case-folded to lowercase; internal whitespace is preserved, and no punctuation removal or tokenization is applied. Let \(\tilde{a}\) and \(\tilde{b}\) denote the normalized strings. We compute text consistency by character-level longest common subsequence:
\begin{equation}
\mathrm{TextSim}(a,b)
=
\begin{cases}
1, & |\tilde{a}|=|\tilde{b}|=0,\\[2pt]
\frac{|\mathrm{LCS}(\tilde{a},\tilde{b})|}
{\max(|\tilde{a}|,|\tilde{b}|)}, & \text{otherwise}.
\end{cases}
\end{equation}
TextSim is used only as a matching criterion for B-SLR, not as the CER computation itself.

\subsection{Canonical Label Mapping}
\label{sec:craw_to_c5_appendix}

\begingroup
\fussy
Table~\ref{tab:craw_to_c5} reports the label normalization used for detection evaluation and category-level diagnostics. 
Raw dataset and parser labels are mapped to five canonical layout categories: text, title, table, figure, and equation. We denote this set by $\mathcal{C}_{5}$; non-content labels are ignored for mAP.
\par
\endgroup

\begin{table*}[t]
\centering
\small
\setlength{\tabcolsep}{4.5pt}
\begin{tabular}{p{0.18\linewidth}p{0.50\linewidth}p{0.14\linewidth}>{\raggedright\arraybackslash}p{0.10\linewidth}}
\toprule
Source family & Raw label(s) & Canonical target & Used in mAP \\
\midrule
PubLayNet GT & Text; List & text & yes \\
PubLayNet GT & Title & title & yes \\
PubLayNet GT & Table & table & yes \\
PubLayNet GT & Figure & figure & yes \\
DocLayNet GT & Caption; Footnote; List-item & text & yes \\
DocLayNet GT & Section-header & title & yes \\
DocLayNet GT & Picture & figure & yes \\
DocLayNet GT & Formula & equation & yes \\
Parser / MinerU output & figure\_caption; table\_caption; reference; list; plain\_text; table\_footnote; formula\_caption; index; normal\_text & text & yes \\
Parser / MinerU output & image & figure & yes \\
Parser / MinerU output & formula; isolate\_formula; embedding; isolated & equation & yes \\
Canonical passthrough & text; title; table; figure; equation & identity & yes \\
Non-content labels & Page-header; Page-footer; header; footer; abandon; seal & None & no \\
Fallback rule & Any unseen label & text (default) & yes \\
\bottomrule
\end{tabular}
\caption{Label normalization from raw dataset/parser labels to the five canonical layout categories used for detection evaluation.}
\label{tab:craw_to_c5}
\end{table*}

\subsection{Terminal Metric Definitions}
\label{sec:terminal_metric_appendix}

We instantiate the terminal component $\Delta$ through OCR and detection degradation:
\begin{equation}
\Delta=\bigl(\Delta_{\mathrm{ocr}},\Delta_{\mathrm{det}}\bigr),
\end{equation}
where $\Delta_{\mathrm{ocr}}$ is measured by CER and $\Delta_{\mathrm{det}}$ by the per-image mAP@0.5 drop.

For the OCR channel, we use character error rate (CER) as the primary terminal judge. 
Because the audit compares clean and perturbed parser outputs, the clean parser text serves as the reference text for each matched element. 
Using the maximum-IoU correspondence defined in Appendix~\ref{sec:matching_appendix}, let $r_e=s(e)$ and $h_e=s(m(e))$. 
The element-wise CER is
\begin{equation}
\mathrm{CER}_{e}(e)=
\begin{cases}
\dfrac{d_{\mathrm{Lev}}(r_e,h_e)}{|r_e|},
& \mathrm{IoU}(e,m(e))>0,\\[5pt]
1, & \text{otherwise,}
\end{cases}
\label{eq:cer_e}
\end{equation}
where the second case represents total text loss when no perturbed element overlaps the clean element. 
Here $d_{\mathrm{Lev}}$ denotes character-level Levenshtein distance after stripping leading/trailing whitespace and case-folding to lowercase.

Let
\begin{equation}
E_{\mathrm{text}}
=
\left\{
e\in E
\;\middle|\;
s(e)\neq\epsilon
\right\}
\label{eq:ematched}
\end{equation}
denote the subset of clean elements with non-empty text. 
\begingroup
\setlength{\abovedisplayskip}{3pt plus 1pt minus 1pt}
\setlength{\belowdisplayskip}{3pt plus 1pt minus 1pt}
\setlength{\abovedisplayshortskip}{2pt plus 1pt minus 1pt}
\setlength{\belowdisplayshortskip}{2pt plus 1pt minus 1pt}
The page-level mean CER is
\begin{equation}
\overline{\mathrm{CER}}
=
\frac{1}{|E_{\mathrm{text}}|}
\sum_{e\in E_{\mathrm{text}}}
\mathrm{CER}_{e}(e),
\label{eq:mean_cer}
\end{equation}
with $\overline{\mathrm{CER}}=1$ when $|E_{\mathrm{text}}|=0$.

For the detection channel, we compute per-image mean Average Precision at IoU threshold $0.5$:
\begin{equation}
\mathrm{mAP}
=
\frac{1}{|\mathcal{C}|}
\sum_{c\in\mathcal{C}}\mathrm{AP}_{c},
\label{eq:map}
\end{equation}
where $\mathcal{C}\subseteq\mathcal{C}_{5}$ is the set of normalized layout classes present in the page-level ground truth. 
The raw mAP is a detection quality score, so we use its clean-to-perturbed drop as the detection degradation signal:
\begin{equation}
\begin{aligned}
\Delta_{\mathrm{det}}
&:=
\Delta\mathrm{mAP}\\
&=
\mathrm{mAP}(\mathcal{Y},E)
-
\mathrm{mAP}(\mathcal{Y},E_{\mathrm{adv}}).
\end{aligned}
\label{eq:delta_map}
\end{equation}
Positive $\Delta\mathrm{mAP}$ indicates detection degradation, while negative values indicate an incidental mAP improvement after perturbation. CER remains the primary terminal judge for OCR-oriented failure propagation.
\par
\endgroup

\subsection{Operational Thresholds}
\label{sec:opconstants}

\begingroup
\makeatletter
\renewcommand\paragraph{\@startsection{paragraph}{4}{\z@}{1.5ex}{-0.8em}{\normalfont\normalsize\bfseries}}
\makeatother
\paragraph{Coverage ratio $\rho$.}
The coverage ratio
\begin{equation}
\rho(e,P)=\frac{|b(e)\cap A(P)|}{|b(e)|}
\end{equation}
measures the fraction of an element box directly covered by the perturbation mask. 
Using the fixed threshold $\eta_{\mathrm{occ}}=0.3$, a structurally failed element with $\rho(e,P)\ge\eta_{\mathrm{occ}}$ is assigned to the direct-occlusion subset $\mathcal{U}_{\mathrm{miss}}$; otherwise, it is assigned to the topology-dominant subset $\mathcal{U}_{\mathrm{topo}}$.

\paragraph{Alignment threshold rationale.}
The IoU threshold $\tau_{\mathrm{iou}}=0.1$ is intentionally permissive because the IoU gate is used for diagnostic alignment rather than detection-quality scoring.
Unlike standard detection evaluation, where IoU thresholds such as $0.5$ determine whether a detection is correct, our matching stage only admits a candidate perturbed element for subsequent identity checking.
A strict IoU gate would prematurely mark spatially displaced but still identifiable counterparts as lost, thereby conflating geometric displacement with structural disappearance.
This follows the spirit of text-aware detection evaluation, where geometric matching can be relaxed when text-level consistency provides the substantive quality signal~\citep{tedeval2019,cleval2020}.
The TextSim threshold $\tau_{\mathrm{text}}=0.5$ requires at least half of the character-level LCS-normalized content to be preserved; below this value, textual identity is treated as structurally lost.
Both thresholds are fixed across all parsers, probes, and policies.
\endgroup

\begin{table*}[!t]
\centering
\footnotesize
\setlength{\tabcolsep}{5.5pt}
\renewcommand{\arraystretch}{1.12}
\newcommand{\konecell}[1]{\cellcolor{red!8}{#1}}

\textbf{(a) Corpus prevalence of short text blocks}\\[2pt]
\begin{tabular*}{\textwidth}{@{\extracolsep{\fill}}lrrrrr@{}}
\toprule
Parser & Clean blocks & $\lvert t\rvert=1$ & $\lvert t\rvert\leq3$ & $\lvert t\rvert\leq5$ & $\lvert t\rvert\leq10$ \\
\midrule
MinerU
& 12{,}117 & \textbf{37 (0.31\%)} & 79 (0.65\%)
& 137 (1.13\%) & 693 (5.72\%) \\
PP-StructureV3
& 16{,}383 & \textbf{213 (1.30\%)} & 692 (4.22\%)
& 923 (5.63\%) & 1{,}602 (9.78\%) \\
\bottomrule
\end{tabular*}

\vspace{5pt}
\textbf{(b) B-SLR--CER alignment under minimum-length gate ablation}\\[2pt]
\begin{tabular*}{\textwidth}{@{\extracolsep{\fill}}lrrrrrrr@{}}
\toprule
\multirow{2}{*}{Parser}
& \multicolumn{5}{c}{Minimum TextSim-gate length $k$}
& \multirow{2}{*}{$\max\lvert\Delta\rvert$}
& \multirow{2}{*}{TOR} \\
\cmidrule(lr){2-6}
& \konecell{$1$} & $2$ & $3$ & $5$ & $10$ & & \\
\midrule
MinerU
& \konecell{\textbf{.727 (.911)}}
& .724 (.911) & .724 (.905) & .722 (.904) & .715 (.904)
& .012 (.007) & .384 (.482) \\
PP-StructureV3
& \konecell{\textbf{.916 (.973)}}
& .917 (.973) & .916 (.973) & .916 (.973) & .916 (.973)
& .001 (.000) & .110 (.459) \\
\bottomrule
\end{tabular*}

\caption{\textbf{Short-block prevalence and TextSim-gate stability.} (a) Counts and corpus percentages by normalized clean-text length over 1{,}000 pages; bold marks the one-character single-error corner case. (b) Each entry reports configuration-level $R^2$ (Spearman $\rho_s$) between Block-level Structural Loss Rate (B-SLR) and Character Error Rate (CER) over 29 configuration means. The highlighted $k=1$ column marks the parameter used in all reported experiments; $\max\lvert\Delta\rvert$ is the largest absolute change from $k=1$ across the sweep. Total Occlusion Ratio (TOR) is threshold-invariant.}
\label{tab:shortblock_distribution}
\label{tab:shortblock_sensitivity}
\end{table*}

\subsection{Short-Block TextSim Stability}
\label{sec:shortblock_sensitivity}

\begingroup
\setlength{\abovedisplayskip}{3pt plus 1pt minus 1pt}
\setlength{\belowdisplayskip}{3pt plus 1pt minus 1pt}
\setlength{\abovedisplayshortskip}{2pt plus 1pt minus 1pt}
\setlength{\belowdisplayshortskip}{2pt plus 1pt minus 1pt}
We analyze the short-text boundary directly. Let $n=\lvert t\rvert$ be the normalized clean-text length of a nonempty block. After one substitution or deletion that reduces the longest common subsequence (LCS) by one character, the denominator in Eq.~\ref{eq:textsim} remains $n$, and hence
\begin{equation}
\mathrm{TextSim}
=\frac{n-1}{n}<0.5
\iff n=1.
\label{eq:shortblock_single_error}
\end{equation}
For $n=2$, TextSim is exactly $0.5$ and therefore passes the stated $\mathrm{TextSim}\ge\tau_{\mathrm{text}}$ gate. More generally, after $r$ LCS-reducing substitutions or deletions,
\begin{equation}
\frac{n-r}{n}<0.5
\iff r>\frac{n}{2}.
\label{eq:shortblock_r_errors}
\end{equation}
Thus, under $\tau_{\mathrm{text}}=0.5$, there is no universal minimum reliable length independent of the number of LCS-reducing errors; the concrete single-error corner case is a one-character block.
\par
\endgroup

\begingroup
\fussy
Table~\ref{tab:shortblock_distribution}(a) counts this case and broader length ranges over the full 1{,}000-page clean corpus. Only 37 MinerU blocks and 213 PP-StructureV3 blocks fall below the candidate single-substitution/deletion boundary $k=2$. We next vary only the minimum length $k\in\{1,2,3,5,10\}$ at which the TextSim gate is applied. Blocks with $n<k$ remain in the B-SLR denominator and retain every geometry-gate failure; only their text gate is disabled. The main experiments use $k=1$. The reconstruction exactly matches the reported B-SLR on all 58{,}000 page--configuration rows before varying $k$.
\par
\endgroup

As Table~\ref{tab:shortblock_sensitivity} shows, disabling the text gate only for one-character blocks ($k=2$) leaves $\rho_s$ unchanged on both parsers and changes $R^2$ by only $-0.0037$ on MinerU and $+0.0001$ on PP-StructureV3. Even $k=10$ changes $(R^2,\rho_s)$ by only $(-0.0118,-0.0069)$ and $(-0.0002,0)$, respectively. One-character instances account for 2.82\% and 0.77\% of all automatically identified text-gate triggers. Because no manual structural-identity labels are assumed, these cases are reported as text-gate triggers rather than verified false structural losses. The analysis therefore supports a local one-character caveat, but not a minimum-length cutoff for aggregate B-SLR/CER alignment.

\subsection{Matching-Threshold Sensitivity}
\label{sec:threshold_sensitivity}

We recompute B-SLR for all nine combinations in a $3\times3$ Phase-1 threshold grid.
The grid uses $\tau_{\mathrm{iou}}\in\{0.05,0.1,0.3\}$ and
$\tau_{\mathrm{text}}\in\{0.3,0.5,0.7\}$.
Following Figure~\ref{fig:footprint_bias}, we average B-SLR and
$\overline{\mathrm{CER}}$ over the same 1{,}000 pages within each of the
29 configurations and evaluate their configuration-level alignment.
Only the B-SLR matching thresholds vary; the CER target and TOR baseline
remain fixed.

\begin{table*}[t]
\centering
\small
\setlength{\tabcolsep}{7pt}
\renewcommand{\arraystretch}{1.08}
\newcommand{\thresholdcell}[1]{\cellcolor{red!8}{#1}}

\textbf{(a) Configuration-level $R^2$}\\[2pt]
\begin{tabular*}{\textwidth}{@{\extracolsep{\fill}}c ccc ccc@{}}
\toprule
\multirow{2}{*}{$\tau_{\mathrm{iou}}$}
& \multicolumn{3}{c}{\textbf{MinerU}: $\tau_{\mathrm{text}}$}
& \multicolumn{3}{c}{\textbf{PP-StructureV3}: $\tau_{\mathrm{text}}$} \\
\cmidrule(lr){2-4}\cmidrule(l){5-7}
& $.3$ & $.5$ & $.7$ & $.3$ & $.5$ & $.7$ \\
\midrule
$.05$ & $.627$ & $.724$ & $.788$ & $.904$ & $.915$ & $.934$ \\
$.10$ & $.639$ & \thresholdcell{\textbf{.727}} & $.790$ & $.907$ & \thresholdcell{\textbf{.916}} & $.935$ \\
$.30$ & $.632$ & $.721$ & $.789$ & $.889$ & $.918$ & $.937$ \\
\bottomrule
\end{tabular*}

\vspace{5pt}
\textbf{(b) Configuration-level Spearman $\rho_s$}\\[2pt]
\begin{tabular*}{\textwidth}{@{\extracolsep{\fill}}c ccc ccc@{}}
\toprule
\multirow{2}{*}{$\tau_{\mathrm{iou}}$}
& \multicolumn{3}{c}{\textbf{MinerU}: $\tau_{\mathrm{text}}$}
& \multicolumn{3}{c}{\textbf{PP-StructureV3}: $\tau_{\mathrm{text}}$} \\
\cmidrule(lr){2-4}\cmidrule(l){5-7}
& $.3$ & $.5$ & $.7$ & $.3$ & $.5$ & $.7$ \\
\midrule
$.05$ & $.859$ & $.910$ & $.928$ & $.986$ & $.971$ & $.969$ \\
$.10$ & $.864$ & \thresholdcell{\textbf{.911}} & $.937$ & $.985$ & \thresholdcell{\textbf{.973}} & $.969$ \\
$.30$ & $.896$ & $.917$ & $.927$ & $.987$ & $.973$ & $.973$ \\
\bottomrule
\end{tabular*}

\caption{\textbf{Matching-threshold sensitivity of B-SLR--CER alignment.} Panels report configuration-level $R^2$ and Spearman $\rho_s$ between Block-level Structural Loss Rate (B-SLR) and Character Error Rate (CER) over 29 configuration means, each averaged over 1{,}000 pages. Highlighted cells mark the adopted setting $(\tau_{\mathrm{iou}},\tau_{\mathrm{text}})=(0.1,0.5)$. Total Occlusion Ratio (TOR) is threshold-invariant, with $R^2=.384/.110$ and $\rho_s=.482/.459$ on MinerU/PP-StructureV3; B-SLR exceeds TOR under all nine threshold pairs.}
\label{tab:threshold_sensitivity}
\end{table*}

Table~\ref{tab:threshold_sensitivity} shows that the main conclusion is invariant throughout the sweep. Across all nine threshold pairs, B-SLR achieves $R^2=.627$--$.790$ and $\rho_s=.859$--$.937$ on MinerU, and $R^2=.889$--$.937$ and $\rho_s=.969$--$.987$ on PP-StructureV3. Even the weakest B-SLR result exceeds the corresponding TOR baseline by $.243/.377$ on MinerU and $.779/.510$ on PP-StructureV3 in $R^2/\rho_s$, respectively.

The alignment is particularly insensitive to $\tau_{\mathrm{iou}}$: at a fixed $\tau_{\mathrm{text}}$, changing $\tau_{\mathrm{iou}}$ changes $R^2$ by at most $.012/.017$ and $\rho_s$ by at most $.038/.004$ on MinerU/PP-StructureV3. The absolute MinerU alignment is more sensitive to the text-similarity gate, but remains strong even at the least favorable setting. The adopted setting $(0.1,0.5)$ is retained rather than selected retrospectively from this sweep and exactly reproduces the Figure~\ref{fig:footprint_bias} results $(.727,.911)$ and $(.916,.973)$.

\subsection{B-SLR Decomposition Algorithm}
\label{sec:bslr_pseudocode}

Algorithm~\ref{alg:bslr} gives the implementation-level procedure used to decompose B-SLR into direct-occlusion and topology-dominant components. 
The algorithm first identifies structurally failed elements using the same IoU--TextSim gates as the main B-SLR definition. 
Failed elements are then split into direct-occlusion and topology-dominant subsets according to the coverage ratio $\rho(e,P)$. 
Within the topology-dominant subset $\mathcal{U}_{\mathrm{topo}}$, failures are further assigned to three mutually exclusive categories under a fixed priority order: \textsc{merge}, \textsc{misclass}, and \textsc{degraded}. 
\textsc{Merge} captures multiple clean elements assigned to the same perturbed prediction; \textsc{misclass} captures retained spatial correspondence with a changed canonical category; and \textsc{degraded} covers the remaining non-occluded failures, including text degradation without direct physical coverage.

\begin{algorithm}[tbp]
\DontPrintSemicolon
\SetAlgoLined
\SetKwInput{KwIn}{Input}
\SetKwInput{KwOut}{Output}
\KwIn{$E,\; E_{\mathrm{adv}},\; A(P)$}
\KwOut{$\mathrm{SLR}_{\mathrm{miss}},\; \mathrm{SLR}_{\mathrm{topo}}$, per-category counts}
\BlankLine
Compute $m(e)=\arg\max_{e'\in E_{\mathrm{adv}}}\mathrm{IoU}(e,e')$ for each $e\in E$\;
Count how many clean elements map to each perturbed prediction\;
\BlankLine
\ForEach{$e \in E$}{
  $\hat{e}\gets m(e)$\;
  $\mathrm{iou}\gets \mathrm{IoU}(e,\hat{e})$\;
  $\hat{s}\gets \mathrm{TextSim}(e,\hat{e})$\;
  $\mathrm{failed}\gets
  (\mathrm{iou}<\tau_{\mathrm{iou}})
  \vee
  (\hat{s}<\tau_{\mathrm{text}})$\;
  \BlankLine
  \uIf{\textbf{not} $\mathrm{failed}$}{
    \textsc{intact}\;
  }
  \uElseIf{$\rho(e,P) \ge \eta_{\mathrm{occ}}$}{
    \textsc{miss}\;
  }
  \uElseIf{multiple clean elements map to $\hat{e}$}{
    \textsc{merge}\;
  }
  \uElseIf{$\mathrm{iou}\ge\tau_{\mathrm{iou}}$ \textbf{and} $c(e) \ne c(\hat{e})$}{
    \textsc{misclass}\;
  }
  \Else{
    \textsc{degraded}\;
  }
}
\BlankLine
$\mathrm{SLR}_{\mathrm{miss}} \leftarrow n_{\textsc{miss}} / |E|$\;
$\mathrm{SLR}_{\mathrm{topo}} \leftarrow
(n_{\textsc{merge}} + n_{\textsc{misclass}} + n_{\textsc{degraded}}) / |E|$\;
\caption{Decomposed B-SLR classification.}
\label{alg:bslr}
\end{algorithm}

\subsection{Policy-Audit Metric Shorthands}
\label{sec:policy_metric_shorthands}

\begingroup
\setlength{\abovedisplayskip}{3pt plus 1pt minus 1pt}
\setlength{\belowdisplayskip}{3pt plus 1pt minus 1pt}
\setlength{\abovedisplayshortskip}{2pt plus 1pt minus 1pt}
\setlength{\belowdisplayshortskip}{2pt plus 1pt minus 1pt}
Table~\ref{tab:policy_main} uses compact metric names for readability. 
The exposure descriptors TOR, BPO, BOC, and EIR follow the definitions in Appendix~\ref{sec:diag_appendix}. 
CER and $\Delta$mAP are the terminal OCR and detection-side degradation signals defined in Appendix~\ref{sec:terminal_metric_appendix}, and B-SLR is the structural loss rate defined in Section~\ref{sec:failure_measurement}. 
In the policy audit, \textit{Miss} and \textit{Topo} denote $\mathrm{SLR}_{\mathrm{miss}}$ and $\mathrm{SLR}_{\mathrm{topo}}$, respectively. 
TopoShare measures the fraction of structural loss explained by topology-dominant failures:
\begin{equation}
\mathrm{TopoShare}
=
\frac{\mathrm{SLR}_{\mathrm{topo}}}
{\mathrm{SLR}_{\mathrm{miss}}+\mathrm{SLR}_{\mathrm{topo}}}.
\end{equation}
Eff-B and Eff-C denote per-footprint efficiency for structural loss and OCR instability:
\begin{equation}
\mathrm{Eff\mbox{-}B}
=
\frac{\mathrm{B\mbox{-}SLR}}{\mathrm{TOR}},
\qquad
\mathrm{Eff\mbox{-}C}
=
\frac{\mathrm{CER}}{\mathrm{TOR}}.
\end{equation}
These efficiency scores are used only as relative indicators under comparable small-footprint regimes; they should be interpreted together with absolute damage and pathway composition.
\par
\endgroup

\section{Experimental Protocol and Configuration Details}
\label{sec:appendix_protocol}

\subsection{Auditing Protocol Overview and Identifier Convention}
\label{sec:config_matrix}
\label{sec:stats_protocol}

Table~\ref{tab:exp_matrix} summarizes the configuration groups used in our experiments.
We use A01--A22 for fixed standard configurations, NT01--NT07 for EIR-targeted configurations, and S01--S13 for randomized sweep configurations.
For compact reporting, each fixed configuration is encoded as
\begin{equation}
\mathsf{P}_{n}\!\cdot\!t[\xi],
\label{eq:config_label}
\end{equation}
where $\mathsf{P}_{n}$ identifies the probe family in Table~\ref{tab:probe_catalog}, $t\in\{\mathtt{a},\mathtt{c},\mathtt{r},\mathtt{b}\}$ denotes placement, and $\xi$ records the dominant varied parameter.
Tables~\ref{tab:config_a}, \ref{tab:config_nt}, and~\ref{tab:config_s} provide the full decoding.

\begin{table}[tbp]
\centering
\small
\setlength{\tabcolsep}{4pt}
\begin{tabularx}{\columnwidth}{@{}lccY@{}}
\toprule
Protocol & Configs & Per config & Goal \\
\midrule
Phase 1 fixed & 29 & 1{,}000 imgs & probe effect profiling \\
Phase 1 sweep & 13 & 1{,}000 imgs & randomized-parameter check \\
Phase 2 policies & 5 policies & 1{,}000 imgs & targeting-policy comparison \\
\bottomrule
\end{tabularx}
\caption{Overview of auditing protocols and configuration groups.}
\label{tab:exp_matrix}
\end{table}

\subsection{Phase 1 Fixed Configurations}
\label{sec:config_a}

Table~\ref{tab:config_a} decodes the 22 standard Phase~1 configurations. 
These configurations are used to profile probe-level effects under controlled parameter and placement choices.

\begin{table}[tbp]
\centering
\scriptsize
\setlength{\tabcolsep}{2pt}
\resizebox{\columnwidth}{!}{%
\begin{tabular}{@{}llll@{}}
\toprule
Label & Probe tuple label & Key parameters & Purpose \\
\midrule
A01 & $\mathsf{P}_{1}\!\cdot\!\mathtt{a}[w1]$       & $w\!=\!1,\;l\!=\!W$         & P1 lower bound \\
A02 & $\mathsf{P}_{1}\!\cdot\!\mathtt{a}[w8]$       & $w\!=\!8,\;l\!=\!W$         & P1 upper bound \\
A03 & $\mathsf{P}_{2}\!\cdot\!\mathtt{a}[w1]$       & $w\!=\!1,\;l\!=\!H$         & P2 lower bound \\
A04 & $\mathsf{P}_{2}\!\cdot\!\mathtt{a}[w8]$       & $w\!=\!8,\;l\!=\!H$         & P2 upper bound \\
A05 & $\mathsf{P}_{3}\!\cdot\!\mathtt{a}[\alpha.3]$ & $r\!=\!60,\;\alpha\!=\!0.3$ & stamp faint \\
A06 & $\mathsf{P}_{3}\!\cdot\!\mathtt{a}[\alpha1]$  & $r\!=\!60,\;\alpha\!=\!1.0$ & stamp opaque \\
A07 & $\mathsf{P}_{4}\!\cdot\!\mathtt{c}[5\%]$      & $a_{\mathrm{area}}\!=\!5\%,\;\beta\!=\!0.3$  & erasure mild \\
A08 & $\mathsf{P}_{4}\!\cdot\!\mathtt{c}[20\%]$     & $a_{\mathrm{area}}\!=\!20\%,\;\beta\!=\!1.0$ & erasure severe \\
A09 & $\mathsf{P}_{5}\!\cdot\!\mathtt{b}[w1]$       & $w\!=\!1,\;l\!=\!0.5W$       & separator thin \\
A10 & $\mathsf{P}_{5}\!\cdot\!\mathtt{b}[w3]$       & $w\!=\!3,\;l\!=\!0.5W$       & separator thick \\
A11 & $\mathsf{P}_{6}\!\cdot\!\mathtt{a}[\alpha.1]$ & $\alpha\!=\!0.1,\;w\!=\!5$   & ghost imperceptible \\
A12 & $\mathsf{P}_{6}\!\cdot\!\mathtt{a}[\alpha.3]$ & $\alpha\!=\!0.3,\;w\!=\!5$   & ghost visible \\
\midrule
\multicolumn{4}{@{}l@{}}{\textit{Placement control groups}} \\
\multicolumn{4}{@{}l@{}}{\textit{(matched probe, varying $\mathcal{T}$):}} \\
A13 & $\mathsf{P}_{1}\!\cdot\!\mathtt{c}[w3]$       & $w\!=\!3,\;l\!=\!W$         & crease on text \\
A14 & $\mathsf{P}_{1}\!\cdot\!\mathtt{r}[w3]$       & $w\!=\!3,\;l\!=\!W$         & crease random \\
A15 & $\mathsf{P}_{3}\!\cdot\!\mathtt{c}[\alpha.5]$ & $r\!=\!60,\;\alpha\!=\!0.5$ & stamp on text \\
A16 & $\mathsf{P}_{3}\!\cdot\!\mathtt{r}[\alpha.5]$ & $r\!=\!60,\;\alpha\!=\!0.5$ & stamp random \\
A17 & $\mathsf{P}_{5}\!\cdot\!\mathtt{c}[w2]$       & $w\!=\!2,\;l\!=\!0.5W$       & sep.\ on text \\
A18 & $\mathsf{P}_{5}\!\cdot\!\mathtt{r}[w2]$       & $w\!=\!2,\;l\!=\!0.5W$       & sep.\ random \\
A19 & $\mathsf{P}_{4}\!\cdot\!\mathtt{b}[20\%]$     & $a_{\mathrm{area}}\!=\!20\%,\;\beta\!=\!1.0$ & erasure bridge \\
A20 & $\mathsf{P}_{5}\!\cdot\!\mathtt{c}[w3]$       & $w\!=\!3,\;l\!=\!0.5W$       & sep.\ extended \\
A21 & $\mathsf{P}_{3}\!\cdot\!\mathtt{a}[\alpha.5]$ & $r\!=\!60,\;\alpha\!=\!0.5$ & stamp anchor \\
A22 & $\mathsf{P}_{1}\!\cdot\!\mathtt{a}[w3]$       & $w\!=\!3,\;l\!=\!W$         & crease anchor \\
\bottomrule
\end{tabular}%
}
\caption{Phase~1 fixed configurations under the A01--A22 identifier convention. Tuple labels decode probe family, placement, and dominant parameter; $W$/$H$ denote page width/height.}
\label{tab:config_a}
\end{table}

\subsection{EIR-Targeted NT Configurations}
\label{sec:config_nt}

The NT series uses iterative multi-stamp placement to reach target element-interference levels.
The target $\tau$ specifies the intended fraction of clean elements whose boxes overlap the probe mask at injection time.

\begin{table}[tbp]
\centering
\small
\setlength{\tabcolsep}{4pt}
\begin{tabularx}{\columnwidth}{@{}llcY@{}}
\toprule
Label & Probe tuple label & Target $\tau$ & Method \\
\midrule
NT01 & $\mathsf{NT}[.05]$ & 0.05 & targeted stamp placement \\
NT02 & $\mathsf{NT}[.10]$ & 0.10 & targeted stamp placement \\
NT03 & $\mathsf{NT}[.20]$ & 0.20 & targeted stamp placement \\
NT04 & $\mathsf{NT}[.35]$ & 0.35 & targeted stamp placement \\
NT05 & $\mathsf{NT}[.50]$ & 0.50 & targeted stamp placement \\
NT06 & $\mathsf{NT}[.70]$ & 0.70 & targeted stamp placement \\
NT07 & $\mathsf{NT}[1.0]$ & 1.00 & targeted stamp placement \\
\bottomrule
\end{tabularx}
\caption{EIR-targeted configurations under the NT01--NT07 identifier convention. The probe family and placement mechanism are fixed; only the target $\tau$ varies.}
\label{tab:config_nt}
\end{table}

\subsection{Randomized Sweep Configurations}
\label{sec:config_s}

\begingroup
\fussy
In addition to the fixed Phase~1 matrix, we run randomized sweeps to test whether structural-exposure signals persist when probe parameters vary stochastically.
Table~\ref{tab:config_s} lists the randomized configurations. 
Paired configurations share sampled geometry, appearance, and behavior parameters while varying only placement.
\par
\endgroup

\begin{table}[tbp]
\centering
\scriptsize
\setlength{\tabcolsep}{2pt}
\resizebox{\columnwidth}{!}{%
\begin{tabular}{@{}lllll@{}}
\toprule
ID & Probe & Random parameters & $\mathcal{T}$ & Notes \\
\midrule
S01 & P1 & $w\!\sim\!U(1,10)$ & anchor & \\
S02 & P2 & $w\!\sim\!U(1,10)$ & anchor & \\
S03 & P3 & $r\!\sim\!U(30,90),\;\alpha\!\sim\!U(0.2,1)$ & anchor & \\
S04 & P4 & $a_{\mathrm{area}}\!\sim\!U(3\%,25\%),\;\beta\!\sim\!U(0.2,1)$ & content & \\
S05 & P5 & $w\!\sim\!U(1,5),\;l\!\sim\!U(0.2W,0.8W)$ & bridge & \\
S06 & P6 & $\alpha\!\sim\!U(0.05,0.4),\;w\!\sim\!U(2,10)$ & anchor & \\
S07 & P7 & $n\!\sim\!U(10,100),\;r\!\sim\!U(1,4)$ & random & \\
S08 & P8 & $r_b\!\sim\!U(30,80),\;\kappa\!\sim\!U(0.1,0.5)$ & anchor & \\
S09 & P9 & $\theta\!\sim\!U(20,70),\;w\!\sim\!U(1,6)$ & anchor & \\
\midrule
S10 & P1 & $w\!\sim\!U(1,10)$ & content & paired w/ S01 \\
S11 & P1 & $w\!\sim\!U(1,10)$ & random & paired w/ S01 \\
S12 & P3 & $r\!\sim\!U(30,90),\;\alpha\!\sim\!U(0.2,1)$ & content & paired w/ S03 \\
S13 & P3 & $r\!\sim\!U(30,90),\;\alpha\!\sim\!U(0.2,1)$ & random & paired w/ S03 \\
\bottomrule
\end{tabular}%
}
\caption{Randomized sweep configurations. Paired configurations share sampled $\mathcal{H}$, $\mathcal{V}$, and $\mathcal{B}$; only $\mathcal{T}$ varies.}
\label{tab:config_s}
\end{table}

\subsection{Inference, Randomness, and Software}
\label{sec:inference_config}

\begin{table}[tbp]
\centering
\small
\setlength{\tabcolsep}{4pt}
\begin{tabularx}{\columnwidth}{@{}lYY@{}}
\toprule
Parameter & LLM policies & VLM policy \\
\midrule
Backend & DeepSeek-Chat & Qwen-VL-Max \\
Temperature & 0.7 & 0.7 \\
Max tokens & API default & 1{,}024 \\
Response format & \texttt{json\_object} & free-form (parsed) \\
Max retries & 3 & 3 \\
Image long edge & --- & 1{,}024\,px \\
JPEG quality & --- & 85 \\
\bottomrule
\end{tabularx}
\caption{LLM/VLM inference parameters.}
\label{tab:inference}
\end{table}

\begingroup
\paragraph{\mbox{Randomness control.}}\mbox{}\par
\noindent All experiments use NumPy's \texttt{default\_rng}, initialized with the base seed 42.
Paired randomized-sweep configurations use deterministic per-image seeds
\par\vspace{2pt}
\noindent\makebox[\linewidth][c]{$(\text{image\_index}+1)\times 10^5 + \text{pair\_id}$}
\par\vspace{2pt}
\noindent so that $\mathcal{H}$, $\mathcal{V}$, and $\mathcal{B}$ are sampled identically across paired placement strategies while only $\mathcal{T}$ varies.
LLM/VLM policies inherit stochasticity from API sampling; all other components are deterministic given the seed.
\par
\endgroup

\Needspace{8\baselineskip}
\begingroup
\fussy
\setlength{\emergencystretch}{2em}
\paragraph{\mbox{Software versions.}}\mbox{}\par
\noindent
The primary evaluation uses MinerU~v2.7.6 with\linebreak DocLayout-YOLO, PP-StructureV3 with PaddleOCR~3.4.1 and PaddlePaddle~3.3.1, Python~3.10, NumPy~2.2.6, and OpenCV~4.x. The supplementary grounded-VLM audit uses PaddleOCR~3.7.0, PaddlePaddle~3.2.1, and the PaddleOCR-VL pipeline v1.6.
\par
\endgroup

\paragraph{Hardware.}
GPU experiments used one NVIDIA GeForce RTX 5090 with 32\,GiB device memory.

\subsection{Statistical Verification Layers}
\label{sec:stats_appendix}

We use the six verification layers in Table~\ref{tab:stat_layers} to check that the main conclusions do not depend on a single aggregation view.
The main text emphasizes configuration-level analyses, while the remaining layers provide robustness checks against document heterogeneity, nonlinearity, and attack-design confounds.

\begin{table}[tbp]
\centering
\scriptsize
\setlength{\tabcolsep}{3pt}
\begin{tabularx}{\columnwidth}{@{}cYYY@{}}
\toprule
Layer & Method & Unit & Threat checked \\
\midrule
0 & Raw OLS & per-image record & baseline association \\
1 & Config-level OLS & 29 config means & within-image noise \\
2 & Image fixed effects & de-meaned residuals & document heterogeneity \\
3 & Per-image Spearman & per-image $\rho$ + win rate & nonlinearity \\
4 & Dose-response bins & EIR quintile bins & reverse causality \\
5 & Quasi-natural & within-config quartiles & attack-design confound \\
\bottomrule
\end{tabularx}
\caption{Six-layer statistical verification framework.}
\label{tab:stat_layers}
\end{table}

%% ================================================================
%%  Appendix C — Additional Results and Robustness Checks
%% ================================================================
\section{Additional Results and Robustness Checks}
\label{sec:appendix_additional_results}

\subsection{Channel-by-Exposure Breakdown}
\label{sec:extended_tables}

Table~\ref{tab:fail_channel_exposure} expands the exposure analysis in Section~\ref{sec:exposure_result} by reporting configuration-level $R^2$ between each diagnostic descriptor and each B-SLR channel. This breakdown complements Table~\ref{tab:exposure_pathway} by showing how the five exposure descriptors relate to composite B-SLR and its IoU-fail and text-fail-only subchannels.

\begin{table}[tbp]
\centering
\small
\setlength{\tabcolsep}{4pt}
\begin{tabular}{l l ccc cc}
\toprule
 & & \multicolumn{3}{c}{$\mathcal{G}_{\text{area}}$} & \multicolumn{2}{c}{$\mathcal{G}_{\text{struct}}$} \\
\cmidrule(lr){3-5}\cmidrule(lr){6-7}
Pipeline & Channel & TOR & ACR & BPO & BOC & EIR \\
\midrule
\multirow{3}{*}{MinerU}
 & composite & .780          & .760 & \textbf{.821} & .336          & .280 \\
 & iou-only  & .782          & .755 & \textbf{.869} & .194          & .149 \\
 & text-only & \textbf{.629} & .622 & .613          & .477          & .424 \\
\addlinespace
\multirow{3}{*}{PP-V3}
 & composite & .150 & .149 & .161          & \textbf{.856} & .852 \\
 & iou-only  & .704 & .678 & \textbf{.781} & .203          & .205 \\
 & text-only & .017 & .018 & .017          & \textbf{.738} & .733 \\
\bottomrule
\end{tabular}
\caption{Channel-by-exposure regression breakdown: configuration-level $R^2$ from univariate OLS fits between each diagnostic descriptor and each B-SLR channel. Bold marks the strongest predictor per row.}
\label{tab:fail_channel_exposure}
\end{table}

\paragraph{Fixed-budget pathway triage.}
As a complementary ranking analysis, we pool the 29{,}000 fixed-configuration and 13{,}000 randomized-sweep page--configuration records for each parser (42{,}000 per parser). For each target, positives are records in its top decile, and Precision@10\% evaluates the top decile ranked by a candidate diagnostic. BPO identifies high SLR$_{\mathrm{miss}}$ with precision .481/.486 on MinerU/PP-V3. For high SLR$_{\mathrm{topo}}$, EIR and BOC reach .271/.213 on MinerU and .524/.534 on PP-V3, compared with .052/.055 for TOR. For high CER, B-SLR reaches .741/.828, compared with .233/.207 for TOR. Thus, the regression-level specialization also holds under a fixed inspection budget and for a metric external to B-SLR.

\subsection{Randomized Sweep Robustness}
\label{sec:random_sweep_results}

The randomized sweep configurations in Appendix~\ref{sec:config_s} test whether the structural-exposure signal persists when probe size and appearance vary stochastically rather than being fixed by the Phase~1 matrix. At the configuration level, EIR remains more predictive of B-SLR than footprint-style descriptors: $R^2(\mathrm{EIR}\!\to\!\mathrm{B\mbox{-}SLR})=0.613$ on MinerU and $0.530$ on PP-V3, while TOR/ACR/BPO reach at most $0.048$ on MinerU and $0.020$ on PP-V3. BOC shows the same trend ($R^2=0.591$ on MinerU and $0.532$ on PP-V3). Thus, the dominance of structure-count descriptors is not an artifact of hand-selected fixed configurations.

\subsection{Transfer to Standard Corruptions}
\label{sec:standard_corruption_transfer}

We test whether the B-SLR--CER relationship is confined to the controlled-probe catalog using 24{,}000 cached page--configuration records from four standard corruption families: Gaussian noise, Gaussian blur, JPEG compression, and brightness shift. Each family is evaluated at three severity levels over the same 1{,}000 pages and both primary parsers, yielding 12 complete configurations and 12{,}000 records per parser. The three severities use Gaussian-noise standard deviations $8/16/24$, Gaussian-blur kernels $3/5/9$, JPEG qualities $70/45/25$, and randomly signed brightness-shift magnitudes $22/45/70$.

For the primary configuration-level comparison, we average B-SLR and perturbation-induced CER over the 1{,}000 pages in each configuration and compute Spearman correlation across the 12 resulting means. Table~\ref{tab:standard_corruption_transfer} also reports the pooled page--configuration correlation as supporting descriptive evidence.

\begin{center}
\begin{minipage}{\columnwidth}
\centering
\small
\setlength{\tabcolsep}{5pt}
\begin{tabular}{lcc}
\toprule
Parser & Configuration-level $\rho_s$ & Pooled $\rho_s$ \\
\midrule
MinerU & \textbf{.979} & .845 \\
PP-StructureV3 & \textbf{.937} & .805 \\
\bottomrule
\end{tabular}
\captionof{table}{\textbf{Transfer to standard corruptions.} Spearman correlation ($\rho_s$) between Block-level Structural Loss Rate (B-SLR) and perturbation-induced Character Error Rate (CER) over 12 corruption--severity configurations, with 1{,}000 pages per configuration. The pooled column contains 12{,}000 page--configuration records per parser.}
\label{tab:standard_corruption_transfer}
\end{minipage}
\end{center}

The strong correlations show that B-SLR tracks CER beyond the controlled-probe catalog. This experiment provides out-of-catalog transfer evidence for the diagnostic relationship; it is not a test of whether the controlled probes reproduce complete physical capture processes.

\subsection{Dose-Response Monotonicity}
\label{sec:dose_response_appendix}

We further check whether structural interference produces a monotonic dose-response in B-SLR. Binning observations by EIR, PP-V3 exhibits strict monotonic growth: mean B-SLR increases from $0.018$ in the lowest bin to $0.480$ in the highest bin. MinerU also increases monotonically, from $0.025$ to $0.181$, but with a shallower slope. Within fixed configurations, partitioning images into EIR quartiles yields monotonically increasing B-SLR from Q1 to Q4 in the majority of configurations for both pipelines. Together with the randomized sweep in Section~\ref{sec:random_sweep_results}, this supports that the structural-exposure effect is not an artifact of either specific configuration choices or a single aggregation view.

\subsection{Grounded VLM Scope Extension Details}
\label{sec:vlm_scope_appendix}

We froze a stratified 80-page subset containing 221 existing QA pairs (seed 42) before running PaddleOCR-VL-1.6. Its \texttt{block\_bbox}, \texttt{block\_label}, and \texttt{block\_content} outputs were mapped to ProSA's box--category--text schema. The mean TORs were $.0935\%$ for area-matched erasure and $.0980\%$ for the structural probe. All 240 clean, area-matched, and structural VLM runs succeeded. Regenerated perturbed inputs matched all 80 frozen SHA-256 hashes in each perturbation condition and reproduced the frozen TOR records. From a fresh pipeline initialization, 10/10 clean reruns exactly reproduced the original serialization, yielding clean-to-repeat B-SLR${=}0$. Table~\ref{tab:vlm_scope_full} reports point estimates; the significance statements below use 95\% confidence intervals from 5{,}000 stratified, page-clustered paired-bootstrap resamples.

\begin{table}[t]
\centering
\scriptsize
\setlength{\tabcolsep}{2.5pt}
\renewcommand{\arraystretch}{1.08}
\begin{tabular}{@{}lrrrr@{}}
\toprule
Parser & $\Delta$B-SLR & \shortstack{$\Delta$ Answer-\\missing rate} & BM25 loss & Dense loss \\
\midrule
MinerU & .116 & .127 & .131 & .122 \\
PP-StructureV3 & .100 & .118 & .127 & .118 \\
PaddleOCR-VL-1.6 & \textbf{.056} & .023 & .014 & .009 \\
\bottomrule
\end{tabular}
\caption{\textbf{Matched-footprint grounded-parser audit} on 80 frozen pages and 221 QA pairs. The $\Delta$B-SLR and $\Delta$ Answer-missing rate columns report Structural$-$area-matched differences; the BM25/Dense columns report Answer Hit@5 loss using area-matched$-$Structural. Positive values indicate greater structural-probe damage. Bold marks the PaddleOCR-VL B-SLR gap whose bootstrap interval excludes zero; B-SLR is interpreted within parser only.}
\label{tab:vlm_scope_full}
\end{table}

PaddleOCR-VL-1.6 shows a significant B-SLR gap, while its answer-missing and retrieval intervals include zero. This supports compatibility of the grounded audit interface, but not downstream replication or comparisons of parser robustness. The prespecified 206-pair common-clean-answer sensitivity analysis preserves these directions.

\subsection{Downstream Propagation Details}
\label{sec:downstream_appendix}

The downstream propagation experiment tests whether parser-level structural loss affects practical document use. We construct a frozen set of 975 clean-derived extractive QA pairs from 349 pages selected from a 500-page candidate subset, with at most four pairs per page. Candidates are generated with \texttt{deepseek-chat} ($T=0.2$) from a shared clean source comprising MinerU lines that also occur in PP-StructureV3's clean text. When this shared source is too short, generation deterministically uses one parser's full clean output, selected by text length and malformed-character ratio. We retain a candidate only if its 2--15-token gold answer occurs verbatim in both parsers' clean outputs, maps to an evidence block, and preserves per-page answer-type diversity. Each retained pair contains a question, gold answer span, evidence text span, answer type, and page identifier.

The final set contains 649 span, 183 title, 123 numeric, and 20 caption questions. It contains no free-form or multi-step reasoning questions; the experiment evaluates evidence preservation rather than general reasoning robustness. Table~\ref{tab:qa_type_gaps} reports the matched-footprint exact-match gap for each type.

\Needspace{13\baselineskip}
\noindent\begin{minipage}{\columnwidth}
\centering
\small
\setlength{\tabcolsep}{6pt}
\begin{tabular}{lrrr}
\toprule
Answer type & Count & MinerU & PP-V3 \\
\midrule
Span & 649 & .072 & .059 \\
Title & 183 & .060 & .027 \\
Numeric & 123 & .065 & .041 \\
Caption & 20 & .050 & .050 \\
\bottomrule
\end{tabular}
\captionof{table}{\textbf{QA type distribution and matched-footprint EM gaps.} Values are AM$-$Str.; positive means lower EM under Str. Counts sum to 975. The caption row ($n=20$) warrants caution.}
\label{tab:qa_type_gaps}
\end{minipage}
\par\vspace{6pt}

\paragraph{Perturbation conditions.}
We compare four conditions: clean pages, area-matched erasure (AM), structural probe (Str.), and large-area erasure (LA). 
AM and Str. have nearly identical footprint, with mean TORs of $0.10\%$ and $0.11\%$, respectively. 
LA serves as a large-footprint control with mean TOR $16.61\%$, roughly $150\times$ larger than the matched-footprint conditions.

\paragraph{Downstream QA.}
For each parser and condition, the parsed page text is serialized as the context for a fixed QA answerer. 
The answerer is instructed to return the shortest extractive span using only the provided parsed text; if the answer is absent, it returns \texttt{NOT\_FOUND}. 
We report exact-match accuracy (EM), token-level F1, answer-missing rate, and the EM drop from the clean condition. 
A QA instance is counted as answer-missing when the gold answer string is absent from the parsed page text. 
Empty parser outputs are retained and counted as incorrect and answer-missing when applicable.

The primary answerer uses the \texttt{deepseek-chat} API alias (served identifier \texttt{deepseek-v4-flash}) with temperature 0, $\mathrm{top\_p}=1$, and a 1{,}024-token limit. We evaluate two additional answerers using the identical extractive prompt, scoring protocol, and deterministic decoding setup.

All three answerers cover all 975 frozen QA pairs under every condition. For the GLM-5.2 rerun, batching by 349 pages, four conditions, and two parsers yielded $349\times4\times2=2{,}792$ page--condition requests. All requests completed successfully, with no API-level failures or missing response records. The positive AM$-$Str. gaps persist across all three answerers and both parsers, suggesting that the observed downstream effect is not specific to DeepSeek and is robust across the tested QA answerers.

\Needspace{15\baselineskip}
\noindent\begin{minipage}{\columnwidth}
\centering
\footnotesize
\setlength{\tabcolsep}{3pt}
\begin{tabular}{@{}lrrr@{}}
\toprule
Answerer & $N$ & MinerU & PP-V3 \\
\midrule
DeepSeek (primary) & 975 & .069/.057 & .050/.048 \\
GLM-4-Plus & 975 & .061/.059 & .046/.050 \\
GLM-5.2 & 975 & .075/.059 & .065/.060 \\
\bottomrule
\end{tabular}
\captionof{table}{\textbf{QA-answerer sensitivity.} DeepSeek denotes the primary \texttt{deepseek-chat} answerer. Values are area-matched-minus-structural (AM$-$Str.) EM/F1 gaps on the same paired QA instances; positive values indicate lower performance under the structural probe.}
\label{tab:qa_answerer_sensitivity}
\end{minipage}
\par\vspace{6pt}

\paragraph{\mbox{Page-internal retrieval.}}
For retrieval, the corpus is restricted to chunks from the QA's own page under the same parser and perturbation condition. 
This page-internal setting aligns retrieval with the single-page QA task and isolates whether the perturbed parser output still contains retrievable evidence. 
We use block-aware chunking with a target length of 400 characters, minimum length 80, maximum length 700, and overlap 80. 
Two retrieval backends are evaluated: BM25 and the dense encoder \texttt{all-MiniLM-L6-v2}. 
For each QA, we retrieve the top 10 chunks.

\paragraph{\mbox{Retrieval metrics.}}
We report two retrieval signals. 
Evidence Recall@$k$ checks whether any top-$k$ chunk contains the evidence text, while Answer Hit@$k$ (abbreviated as AnsHit@$k$ in tables) checks whether any top-$k$ chunk contains the gold answer. 
The main text reports Answer Hit@5 because it directly measures whether the answer-bearing content remains retrievable. 
We additionally report evidence Recall@5 and evidence MRR@10 in Table~\ref{tab:downstream_retrieval_full}.

\begin{table*}[tbp]
\centering
\scriptsize
\setlength{\tabcolsep}{4pt}
\renewcommand{\arraystretch}{0.95}
\begin{tabular}{llrrrrrr}
\toprule
Parser & Condition 
& BM25 R@5$_{\mathrm{evid}}$ 
& BM25 MRR@10$_{\mathrm{evid}}$ 
& BM25 AnsHit@5 
& Dense R@5$_{\mathrm{evid}}$ 
& Dense MRR@10$_{\mathrm{evid}}$ 
& Dense AnsHit@5 \\
\midrule
\multirow{4}{*}{MinerU}
& Clean & 68.7 & 61.6 & 94.6 & 67.9 & 56.9 & 93.6 \\
& AM    & 66.7 & 59.6 & 93.8 & 65.7 & 55.2 & 92.8 \\
& Str.  & 49.2 & 43.3 & 81.0 & 48.4 & 41.2 & 80.4 \\
& LA    & 35.6 & 31.2 & 58.8 & 35.4 & 28.7 & 59.4 \\
\midrule
\multirow{4}{*}{PP-V3}
& Clean & 66.4 & 59.3 & 94.7 & 64.5 & 53.1 & 95.1 \\
& AM    & 63.9 & 56.9 & 92.9 & 62.2 & 51.2 & 93.4 \\
& Str.  & 52.7 & 46.8 & 83.1 & 51.5 & 41.5 & 83.5 \\
& LA    & 32.6 & 29.2 & 57.9 & 31.9 & 26.8 & 58.7 \\
\bottomrule
\end{tabular}
\caption{Additional page-internal retrieval metrics for downstream propagation. 
All scores are percentages and higher is better. R@5$_{\mathrm{evid}}$ and MRR@10$_{\mathrm{evid}}$ use evidence-text containment as the relevance signal, while AnsHit@5 checks whether a top-5 retrieved chunk contains the gold answer.}
\label{tab:downstream_retrieval_full}
\end{table*}

\begin{figure*}[t]
\centering
\includegraphics[width=\linewidth]{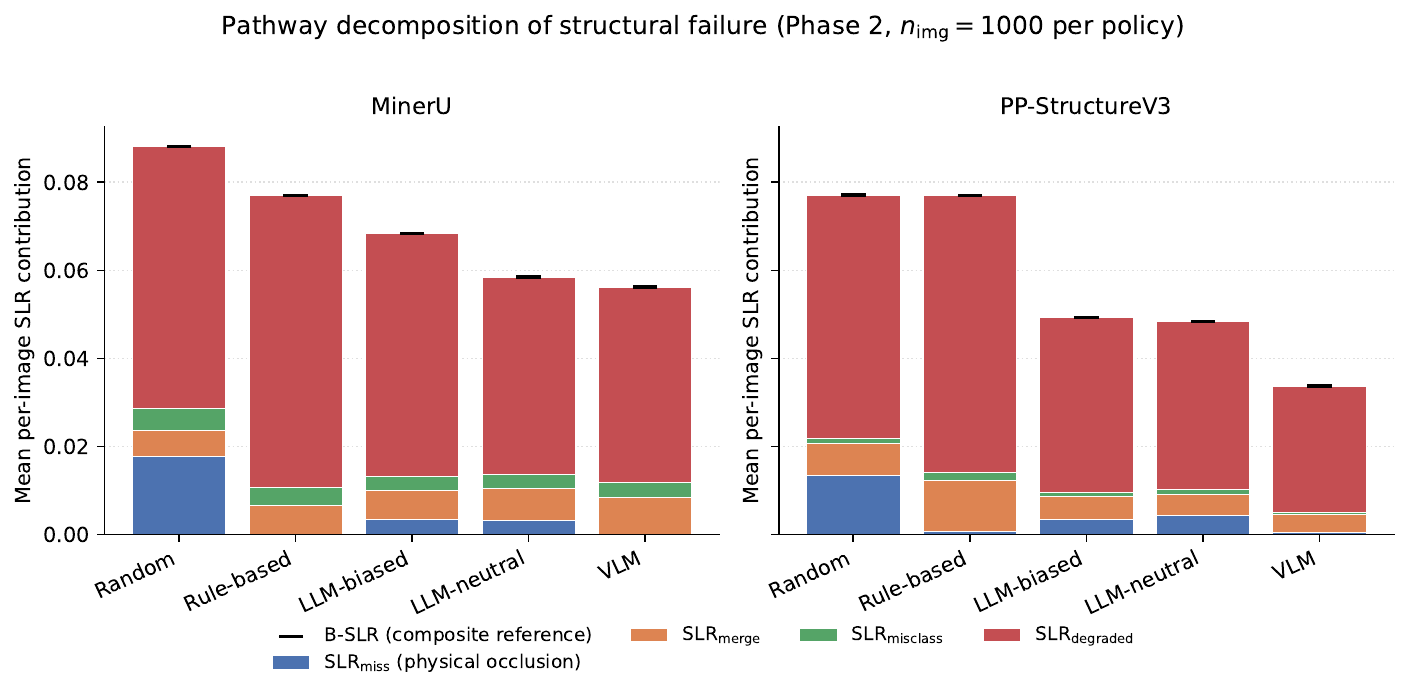}
\caption{Phase~2 pathway composition of mean per-image structural loss for each policy ($n{=}1{,}000$ pages). Stacks show SLR$_{\mathrm{miss}}$, SLR$_{\mathrm{merge}}$, SLR$_{\mathrm{misclass}}$, and SLR$_{\mathrm{degraded}}$; black ticks mark composite B-SLR, and higher stacks indicate greater structural loss.}
\label{fig:pathway_composition}
\end{figure*}

The full retrieval metrics are consistent with the main-text Answer Hit@5 results. In the main-text table, BM25 A@5 and Dense A@5 refer to Answer Hit@5, i.e., whether any top-5 retrieved chunk contains the gold answer string. Under matched footprint, the structural probe causes much larger retrieval degradation than area-matched erasure across both parsers and both retrieval backends. 
For example, on MinerU, BM25 evidence R@5 drops from $66.7\%$ under AM to $49.2\%$ under Str., while Answer Hit@5 drops from $93.8\%$ to $81.0\%$. 
On PP-V3, the same pattern holds: BM25 evidence R@5 drops from $63.9\%$ to $52.7\%$, and Answer Hit@5 drops from $92.9\%$ to $83.1\%$. 
Thus, the downstream effect is not specific to the QA answerer or to a single retrieval backend.

\subsection{Policy Pathway Composition}
\label{sec:policy_pathway_appendix}

Figure~\ref{fig:pathway_composition} provides the fine-grained pathway composition behind the Phase~2 policy comparison in Table~\ref{tab:policy_main}. Across policies, topology-dominant components account for most structural loss, while direct-occlusion loss remains small for targeted policies. This corroborates the main-text observation that efficient targeting exposes topology-level vulnerability rather than simply increasing physical coverage.

\begin{figure*}[t]
\centering
\includegraphics[width=\textwidth]{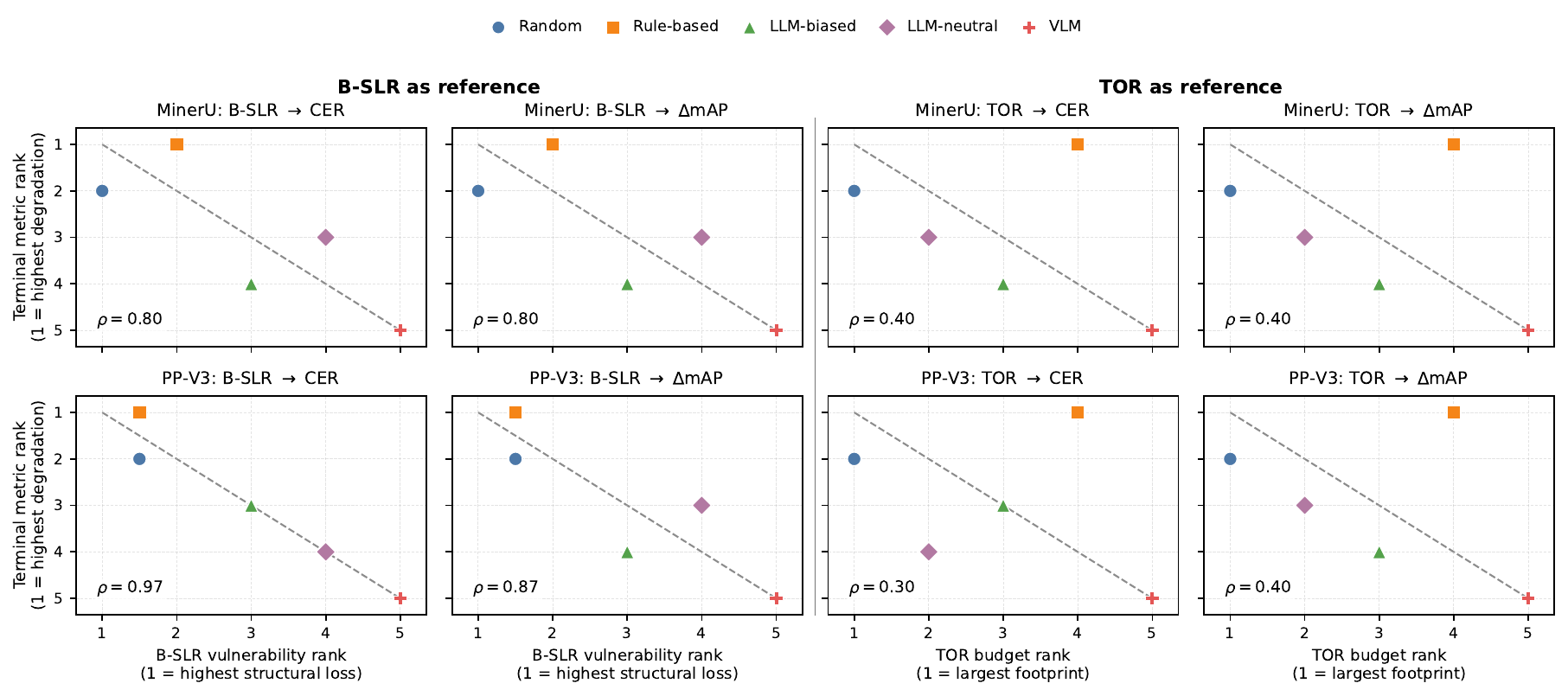}
\caption{\textbf{Policy-level rank consistency.}
Ranks are computed within each parser over the five Phase~2 policies, where rank 1 denotes the highest degradation or largest footprint.
Compared with TOR, B-SLR gives policy rankings that align more closely with CER and \(\Delta\)mAP.}
\label{fig:policy_rank_consistency}
\end{figure*}

\subsection{Policy-Level Rank Consistency}
\label{sec:policy_rank_consistency}

We further examine whether the policy-level ordering induced by structural damage is consistent with terminal degradation signals.
For each parser, we rank the five Phase~2 policies by B-SLR, CER, and \(\Delta\)mAP, where rank 1 denotes the most severe degradation.
As a footprint-based baseline, we also rank policies by TOR and compare the induced ordering with the same terminal metrics.
Figure~\ref{fig:policy_rank_consistency} shows that B-SLR yields a policy ordering that is more consistent with CER and \(\Delta\)mAP than TOR.
The B-SLR-based rank correlations are \(0.80/0.80\) on MinerU and \(0.97/0.87\) on PP-StructureV3 for CER/\(\Delta\)mAP, whereas the corresponding TOR-based correlations drop to \(0.40/0.40\) and \(0.30/0.40\).
This supports the use of B-SLR as a policy-level structural diagnostic: it preserves the relative vulnerability pattern reflected by terminal quality signals better than the area-based footprint proxy.

%% ================================================================
%%  Appendix D — Reproducibility, Prompts, and Scope
%% ================================================================
\section{Prompts, Reproducibility, and Scope}
\label{sec:appendix_repro_scope}

\subsection{Reproducibility Checklist}
\label{sec:checklist}

\begin{itemize}
    \setlength{\itemsep}{2pt}
    \setlength{\parsep}{0pt}
    \setlength{\parskip}{0pt}
    \item \textbf{Data}: PubLayNet and DocLayNet validation splits; pages with fewer than five original spans are filtered; the final shared evaluation pool contains 1{,}000 pages.
    \item \begingroup\fussy\setlength{\emergencystretch}{1em} \textbf{Code}: The official implementation is available at the following URL:\par\noindent\url{https://github.com/ef1026/ProSA}\par\noindent The repository includes a lightweight CPU-only audit module with author-written synthetic examples and unit tests, the main experiment code, adapters for the two primary parsers, frozen attack plans and policy-replay files, an ID-only evaluation manifest, and the downstream QA/retrieval pipelines. To respect third-party licenses and terms, it does not redistribute source page images or annotations, extracted document text, cached parser outputs, QA content, model weights, API responses, or generated reports. The core audit can be run directly from the synthetic example; paper-scale reproduction requires users to obtain the corresponding datasets, parser environments, model weights, and API services.\par\endgroup
    \item \textbf{Pipelines}: The 1{,}000-page primary evaluation uses MinerU~v2.7.6 with DocLayout-YOLO and PP-StructureV3 with RT-DETR-H and PaddleOCR~3.4.1; the supplementary 80-page audit uses PaddleOCR-VL-1.6. Software and hardware details are reported in Appendix~\ref{sec:inference_config}.
    \item \textbf{Matching}: maximum-IoU lookup with unified IoU and TextSim gates ($\tau_{\mathrm{iou}}\!=\!0.1$, $\tau_{\mathrm{text}}\!=\!0.5$); no one-to-one Hungarian assignment is enforced.
    \item \textbf{Probes}: P1--P9 with bounded configuration ranges, listed in Table~\ref{tab:probe_catalog}.
    \item \textbf{Configurations}: A01--A22 and NT01--NT07 for Phase~1 fixed auditing (Tables~\ref{tab:config_a} and~\ref{tab:config_nt}); S01--S13 for randomized sweeps (Table~\ref{tab:config_s}); five policy families for Phase~2.
    \item \textbf{Statistics}: six-layer verification protocol, including configuration-level OLS, image fixed effects, per-image rank checks, dose-response bins, and within-configuration quartile checks (Table~\ref{tab:stat_layers}).
    \item \textbf{Randomness}: base seed 42 for deterministic components; paired randomized-sweep configurations use deterministic per-image seeds; LLM/VLM policies inherit stochasticity from API sampling.
    \item \textbf{Output fields}: Paper-scale experiment outputs record image and configuration identifiers; TOR, ACR, BPO, BOC, and EIR; composite and channel-wise B-SLR; matched CER; clean and perturbed mAP; $\Delta$mAP; and the original-span count. The lightweight audit module separately exposes B-SLR, CER, and the direct-occlusion/topology pathway split for user-provided clean and perturbed outputs.
    \item \textbf{Metrics}: $\overline{\mathrm{CER}}$ is the primary terminal OCR judge; per-image mAP@0.5 over the five canonical layout classes defines the supplementary detection-channel drop $\Delta\mathrm{mAP}$.
\end{itemize}

\Needspace{16\baselineskip}
\subsection{Prompt Templates}
\label{sec:prompts_appendix}

\begingroup
\fussy
\setlength{\emergencystretch}{1em}
All three prompt-based policies emit a probe type, bounded parameters, and a placement strategy. The biased LLM uses a structure-aware renderer, whereas the neutral LLM uses a coordinate-only renderer; both summarize visual, layout, and spatial cues. The VLM instead receives the raw page image. The neutral and VLM policies use these internal strategy mappings: \texttt{between} maps to \texttt{bridge}, \texttt{edge} to \texttt{anchor}, \texttt{inside} to \texttt{content}, and \texttt{anywhere} to \texttt{random}.
The controlled excerpts are shown below.
\par
\endgroup

\smallskip
\begingroup
\setlength{\fboxsep}{4pt}
\setlength{\fboxrule}{0.4pt}
\noindent\fbox{%
\begin{minipage}{\dimexpr\linewidth-2\fboxsep-2\fboxrule\relax}
\small
\textbf{LLM-biased prompt excerpt.}
Role: expert adversarial tester for DLA. Goal: choose one visual probe and placement to maximize parsing failure, including merge, split, or missing-block errors.
Probes: P1--P9; strategies: \texttt{bridge}/\texttt{anchor}/\texttt{content}/\texttt{random}; output: valid JSON.
\par
\textit{Context:} structure-aware page description with block coordinates, gap information, and candidate vulnerable regions.
\end{minipage}}
\par\vspace{5pt}
\noindent\fbox{%
\begin{minipage}{\dimexpr\linewidth-2\fboxsep-2\fboxrule\relax}
\small
\textbf{LLM-neutral prompt excerpt.}
Role: quality-assurance analyst for DLA robustness. Goal: select one realistic perturbation probe and placement strategy to test model response.
Strategies: \texttt{between}/\texttt{edge}/\texttt{inside}/\texttt{anywhere}; output: valid JSON.
\par
\textit{Context:} coordinate-only page description with block types and bounding boxes, without explicit structural vulnerability hints.
\end{minipage}}
\par\vspace{5pt}
\noindent\fbox{%
\begin{minipage}{\dimexpr\linewidth-2\fboxsep-2\fboxrule\relax}
\small
\textbf{VLM strategy-only prompt excerpt.}
Input: raw page image only. Goal: inspect visible text blocks, figures, tables, and whitespace, then select a probe type and placement strategy.
Exact coordinates are determined automatically from the chosen strategy; output: valid JSON.
\end{minipage}}
\endgroup
\smallskip

Invalid strategies default to \texttt{random}, and invalid probe types default to \texttt{P5}.

\end{document}